\documentclass[oupdraft]{bio}
\usepackage[colorlinks=true, urlcolor=citecolor, linkcolor=citecolor, citecolor=citecolor]{hyperref}
\newtheorem{proposition}{Proposition}

\def\var{\rm var}
\usepackage{setspace}

\begin{document}

% Title of paper
\title{Markov Random Fields and Mass Spectra Discrimination}

% List of authors, with corresponding author marked by asterisk
\author{AO KONG\\[4pt]
% Author addresses
\textit{School of Finance, Nanjing University of Finance and Economics, Nanjing, China}
\\[2pt]
 ROBERT AZENCOTT$^\ast$\\[4pt]
 \textit{Dept of Mathematics, University of Houston, TX, USA\\
 Emeritus Prof, C.M.L.A. Ecole Normale Superieure, Cachan, France}
\\[2pt]
% E-mail address for correspondence
{razencot@math.uh.edu}}

\maketitle

% Add a footnote for the corresponding author if one has been
% identified in the author list
\footnotetext{To whom correspondence should be addressed.}

\begin{abstract}
{For mass spectra acquired from cancer patients by MALDI or SELDI techniques, automated discrimination between cancer types or stages has often been implemented by machine learnings. These techniques typically generate ``black-box'' classifiers,   which are difficult to interpret biologically. We develop new and efficient signature discovery algorithms leading to  interpretable signatures combining the discriminating power of explicitly selected small groups of biomarkers, identified by their m/z ratios.  Our approach is based on rigorous stochastic modeling   of ``homogeneous" datasets of mass spectra by a versatile class of parameterized Markov Random Fields. We  present detailed algorithms validated by precise  theoretical results. We also outline the successful tests of our approach to generate efficient explicit signatures for  six benchmark discrimination tasks, based on mass spectra acquired from colorectal cancer patients, as well as from  ovarian cancer patients.}
{MALDI/SELDI data; ovarian cancer; colorectal cancer; biomarker signature discovery; Markov Random Fields; autologistic distribution.}
\end{abstract}

\section{Introduction}
In proteomics, mass spectrometry is a broadly used protein profiling technology to study the mixture of proteins/peptides present in biological tissues or fluids, and is  an  efficient tool for  identification of cancer type and stage (\cite{Eckel2009}).

Mass spectrometry can involve two soft ionization techniques: matrix-assisted laser desorption ionization (MALDI) and surface-enhanced laser desorption and ionization (SELDI). For each analyzed fluid  sample, MALDI or SELDI hardwares generate a high-dimensional mass spectrum, recording between 10,000  and 20,000  ``mass-to-charge (m/z) ratios" corresponding to  the ionized peptides present in the fluid sample,  as well as ``intensities" roughly quantifying the concentrations of these peptides in the sample. Generally m/z ratios take values anywhere between 200 and 20,000 Daltons, and are acquired with a known \textit{relative} accuracy $\rho$ which depends on the acquisition modalities, and ranges from  0.1\% to 0.3\%. 

Analyzing this type of high dimensional data oftern requires specialized software tools, implementing sophisticated machine learning techniques  such as SVM (support vector machines) (\cite{Li2004}, \cite{Yu2005}), artificial neural networks (\cite{Ball2002}), or random forests (\cite{Izmirlian2004}). These techniques typically generate ``black-box'' classifiers, which often reach  good discrimination levels between cancerous   and control groups, but are  difficult to interpret biologically in terms of characteristic biomarkers patterns. This often leads to unexpected performance variations on totally new data sets. To develop clinically usable software tools for analysis of mass spectra acuired by MALDI or SELDI hardwares, a key step is to implement automated discovery of explicit  ``signatures", i.e. short lists of proteomic biomarkers with high discriminating powers between cancer groups (\cite{Yasui2003}). Some easily interpretable automatic classifiers, such as linear combinations of biomarker weights (\cite{Wang2011}), can be found in previous studies, but these approaches do not attempt to quantify the discriminating impact of simultaneous presence for specific pairs of biomarkers.

In this paper, we generate easily interpretable biomarker signatures discriminating between  two arbitrary but homogeneous groups of mass spectra $G^+$ and $G^-$ by  stochastic modeling of biomarkers interactions, taking precisely into account the co-activity of pairs of  biomarkers. To this end,  we fit parametric Markov Random Fields (MRFs) $\pi^+$ and $\pi^-$ to $G^+$ and $G^-$ by Maximum Pseudo-Likelihood Estimation (MPLE). Recall that MRFs have been  successfully used to model spatial dependencies in  high dimensional interacting systems as well as in image and signal analysis (\cite{Chalmond2003}, \cite{Bremaud1999}, \cite{Azencott1992(2)}). Discrimination between $G^+$ and $G^-$  is then achieved by computing the optimal separator between the probability distributions $\pi^+$ and $\pi^-$.

We have studied quite precisely the asymptotic performance of our approach for large data samples, and we have successfully benchmarked  our MRF based signature discovery  technique on MALDI and  SELDI data sets respectively acquired from colorectal and ovarian cancer patients.

\section{Benchmark Mass Spectra Datasets} \label{data}
\subsection{Ovarian Cancer Data ``4-3-02" and ``8-07-02":} 
These two mass spectra data sets, acquired by SELDI-TOF techniques have been previously pre-processed and studied by other authors in \cite{Assareh2007}, \cite{Zhu2003}, \cite{Alexe2004} and can be freely downloaded from the NCI-FDA clinical proteomics databank (http://home.ccr.cancer.gov/ncifdaproteomics/ ppatterns.asp). 

The ``4-3-02" set includes a Control group (CTR) of 116 mass spectra  and an Ovarian Cancer (OVC) group of 100 mass spectra. The ``8-07-02" set includes a Control group (CTR) of 91 mass spectra  and an Ovarian Cancer (OVC) group of 162 mass spectra. As available online, these mass spectra  are already ``aligned" to 15,154 reference m/z ratios ranging from  0 to 20,000 Daltons, and the relative accuracy of m/z ratios is $\rho=0.1\%$. 
\subsection{Colorectal Cancer Data:} For this newly published data set in our previous study \cite{Kong2014}, plasma samples from colorectal cancer patients and a control group were provided by 1st Surgical Clinic, Dept of Surgical, Oncological and Gastroenterological Sciences at University of Padova (Italy). The mass spectra were then acquired through MALDI-TOF techniques by A. Bouamrani, E. Tasciotti, M. Ferrari (Dept. of Nanomedicine, The Methodist Hospital Research Institute, Houston, USA). 

This set includes an Adenoma group (ADE) of 54 spectra, an Early Colorectal Cancer group (ECR) of 80 spectra, a Late Colorectal Cancer group (LCR) of 74 spectra  and a Control group (CTR) of 30 spectra. The union of ADE, ECR, LCR will be called the Colorectal Cancer (CRC) Group. All these mass spectra were generated by MALDI-TOF technique, with a relative accuracy of $\rho = 0.3\%$ on m/z ratios, which range from 800 to 10,000 Daltons.

\subsection{Discrimination Tasks}
Our  paper presents new signature discovery algorithms based on Markov Random Fields modeling of mass spectra. We have then applied and evaluated our MRF discrimination  techniques to the differentiation  between  cancer stages, as well as between cancer and control.  So we have tested implementation and performances of our MRF approach on 6 benchmark discrimination tasks,  using the three sets just described of mass spectra acquired from cancer patients :\\ 
1) ADE vs ECR, ADE vs LCR, ECR vs LCR, CRC vs CTR for the  colorectal cancer dataset, \\ 
2) OVC vs CTR for each one of the two ovarian cancer datasets.

\section{Binary Coding of Mass Spectra}

\subsection{Pre-processing of Proteomic Mass Spectra} \label{preproc}
Pre-processing is an important procedure to lower mass spectra dimensionality and to remove acquisition noise, which affects both m/z ratios and intensities. When an intensity peak is detected at abscissa $A$, its true m/z ratio could lie anywhere within $[ \, A (1-\rho), A (1+ \rho) \, ]$, where the relative accuracy $\rho$ is determined by the acquisition hardware.

For better context control, we apply to each raw mass spectrum our own sequence of pipelined classical pre-processing steps: normalization, smoothing, noise extraction, baseline removal, peak detection as outlined in our previous paper (\cite{Kong2014}). On each mass spectrum, these pre-processing steps detect a usually long list of ``strong intensity peaks''. The m/z abscissas of these detected peaks indicate peptides which could potentially be biomarkers strongly discriminating between cancer types or stages . 

To condense all these approximate peak abscissas, we generate a fixed list of ``reference biomarkers'' $B_s$ with m/z abscissas $B_s = B_1 (1+\rho)^{s-1}, 1 \leq s\leq L$, where $B_1$ and $B_L$ are the smallest and the largest m/z ratios among all spectra in our dataset. Then $S= \{ \, 1, \ldots , L \, \}$ will be called our ``set of sites", and $\mathcal{B} = \mathcal{B}(L)$ will be the set of all binary vectors $x$ of length $L$, with coordinates $x_s$ indexed by $S$. 

\subsection{Binary Coding of Mass Spectra }\label{bincoding}
We say that a reference biomarker $B_s$ is ``activated'' by a mass spectrum $M$ if and only if at least one detected peak of $M$ is positioned within the window of $[\;B_s- \rho B_s,\; B_s+ \rho B_s\;]$. Each mass spectrum $M$ can then be ``coded" by a binary vector $x = x(M) \in \mathcal{B}(L)$ as in \cite{Kong2014}: for each site $s \in S$, we set $x_s(M)= 1 $ if $B_s$ is activated by $M$ and $x_s(M)=0$ otherwise. 

Any group $G$ of $n$ mass spectra thus generates the set $BinG $ of $n$ binary vectors $x(M) \in \mathcal{B}(L)$ coding as just indicated  the mass spectra $M$ of $G$. 

\section{Binary Markov Random Fields and Autologistic Distributions}\label{models}
We now want to consider the data set $BinG $ as a sample of $n$ independent observations of a \textit{random} binary vector $X$ taking values in the set of binary vectors $\mathcal{B}= \mathcal{B}(L)$.

We will systematically model the unknown probability distribution $P$ of $X$ by a Markov Random Field on $\mathcal{B}$. A first statistical  analysis of the data set $BinG$ outlined in  Section \ref{clique} below  will identify for each site $s \in S$ the subset $\mathcal{N}_s$ of all sites $t \in S-s$ for which we ``expect" the coordinate pairs $(X_s, X_t)$ to be strongly correlated.

Recall that a binary random vector $X=\{ X_s\}_{s\in S} $ with values in $\mathcal{B}$ is called a Markov Random Field (MRF) with respect to the family $\mathcal{N}_s$ if for all sites $s\in S$
\[P\{ X_s \mid X_{S\backslash s} \} = P\{ X_s \mid X_{\mathcal{N}_s} \},\]
where for any $K\subset S$, we denote by $X_K$ the set of random variables  $ \{X_s\}_{s\in K}$.
The distribution $P$ of $X$ then belongs to the family of Gibbs distributions, and can be described concretely through its system of ``cliques''. Recall that a clique is a subset $C$ of $S$, such that $t \in \mathcal{N}_s$ for all distinct pairs $\{s, t\} \in C$. All single sites are then cliques of cardinal 1.

Here, we focus on \textit{autologistic} distributions, which are the Gibbs distributions for which all   cliques have cardinal $\leq 2$. They are naturally parameterized by a vector space $\Theta$ isomorphic to $ R^k$, with $k = L(1+L)/2$, and where the coordinates of any parameter vector $\theta \in \Theta $ are denoted by $\theta_s$ and $\theta_{s, t}$ with $s, t \in S$ and $s < t$. 

For each $x\in \mathcal{B}$, let $U(x) \in \Theta$ be the vector with coordinates
\[ U_s(x) = x_s \;, \quad U_{s,t}(x) = x_s x_t \;, \quad \textit{for} \quad s, t \in S , \; s < t. \]
The scalar product of $\theta $ and $ U(x) $ in $\Theta$ is then 
\[
< \theta, U(x) > = \sum_{s\in S} \theta_s x_s +\sum_{s<t}\theta_{s, t}x_s x_t.
\]
For each $\theta \in \Theta$, the autologistic distribution $\pi_{\theta}$ is defined for all $x\in \mathcal{B}$, by
\begin{equation}\label{autologistic}
\pi_{\theta}(x)= \frac{1}{Z(\theta)} e^{- < \theta, U(x) >}, 
\end{equation} 
where $Z(\theta) = \sum_{y\in\mathcal{B}}e^{- < \theta, U(y) >}$ is the partition function.

\section{Fitting Autologistic Distributions to Mass Spectra data sets: Maximum Pseudo-Likelihood Estimators (MPLE)}\label{MPLE}

As seen in Section \ref{bincoding}, the binary coding  of a mass spectra data set $G$  generates a set $BinG$ of  $n$ binary vectors of length $L$, and in concrete applications to cancer data,  $L$ is typically much larger that $n$. Reliable fitting of an autologistic distribution $\pi_{\theta}$ to $BinG$ then requires a strong  dimension reduction from $L$ sites  to a much smaller set $S(d)$ of $d$ sites adequatedly selected in $S$. This is achieved by the  ``Feature Selection'' algorithm we present further on  in Section \ref{feature}. Restricting each binary vector $ x \in BinG$ to the $d$ sites in $S(d)$  transforms $BinG$ into  a set $BG$ of $n$ binary vectors of length $d$.

Defining an autologistic distribution on $\mathcal{B}(d)$ involves selecting a specific  family  of   pairs of sites $\{s, t\}$ with $s,t \in S(d)$, for which the binary  random variables $X_s, X_t$ are expected to have sizeable correlation. The maximum number $d(d-1)/2$ of these potential cliques of order 2 is often still too high with respect to $n$. So we seek model robustness by enforcing parameter parsimony, which leads us to retain only a moderate number $c << d(d-1)/2$ of pertinent cliques of order 2. This is implemented by the ``Clique Discovery'' procedure outlined in Section \ref{clique}. 

After selecting  $d$ sites and a set $\mathcal{C}$ of $c$ cliques of order 2 in $S(d)$,  we seek to model  the data set $BG \subset \mathcal{B}(d)$ of $n$ binary vectors  by an autologistic probability distribution $\pi_{\theta}$ of the form \eqref{autologistic}, where we now impose on $\theta$ the constraints $\theta_s = 0$ whenever $s$ is not in $S(d)$ and $\theta_{s,t} = 0$ whenever $\{s, t\}$ is not in $\mathcal{C}$, so that the unknown parameter vector $\theta $ is now forced to belong to a precise  vector subspace of $\Theta$, of  dimension $(c+d)$.

To achieve this model fitting to data, $\theta$ must be estimated from the $n$ data. After comparative testing of   several classical estimation techniques  on our benchmark examples, we have implemented all our model fitting to data through  Maximum Pseudo-Likelihood Estimators (MPLEs).

These estimators were introduced by \cite{Besag1975} and have played an important role in parameter estimation of spatial models  before the current intensive use  of Monte Carlo methods. Indeed computing the MPLE  requires no simulation of random Gibbs configurations, leading to fast computing speed. 

\subsection{Pseudo-Likelihood }\label{pseudo}
As just seen, the preceding selection of $d$ sites $s$ and $c$ cliques $\{s, t\}$ of order 2 forces a precise set of coordinates of $\theta \in \Theta$ to be equal to $0$. The estimation of the non zero coordinates of $\theta$ will be  based on maximizing the average pseudo-likelihood of the observed data. We now recall how one computes pseudo-likelihoods. For brevity and to simplify  notations, we deliberately restrict our theoretical presentation to the case where no constraints are imposed on the coordinates $\theta_s , \theta_{s,t} $ of $\theta$. The constrained case is an easy extension of the non-constrained case.

Let $x$ be  any observed binary vector and  let $T \in \Theta$ be any tentative estimate of $\theta$. The \textit{pseudo-likelihood} $PL( x, T)$ is classically defined as the product of all ``local specifications" under the tentative autologistic distribution $\pi_T$
\[
PL( x, T)=\prod_s \; \pi_{T}(x_s |  x_{S-s}) = \prod_s \; P_T (Y_s=x_s \mid Y_{S-s}= x_{S-s}),\]
where $Y$ is a random binary  configuration with distribution $\pi_T$.
Consider the linear functions of $T \in \Theta$ defined by    
\begin{equation} \label{aA}
a_{s,x}(T) =  - \, ( T_s+\sum_{s < r} T_{sr} \, x_r ) = \; < A(s,x),T >,
\end{equation}
where  the vector $A(s,x) \in \Theta$ has coordinates
\begin{equation} \label{Asx}
A_s(s,x) = - 1, \quad  A_{sr}(s,x) = -x_r, \quad \textit{for all} \quad s < r.
\end{equation} 
Define two functions of $z$ in $R$ by
\begin{equation} \label{g&h}
g(z)= 1/(1+ e^z), \quad h(z) = e^z/(1+e^z).
\end{equation} 
The conditional specification of $Y_s$ under $\pi_T$ can then be written 
\[
p_{s,x}(T) = P_T ( Y_s = x_s \mid Y_{S-s}= x_{S-s} ) = x_s \, h(a_{s,x}(T)) + (1-x_s) \, g(a_{s,x}(T)). 
\]
Since $x_s$ is either 0 or 1, this easily  implies
\begin{equation} \label{LPLsxT}
LPL_{s,x}(T)= \log( p_{s,x}(T) ) = x_s \, a_{s,x}(T) + \log (\, g(a_{s,x}(T))\, ), 
\end{equation} 
the log pseudo-likelihood function LPL is hence given by 
\[
LPL(x,T) = \sum_{s \in S} LPL_{s,x}(T) =  \sum_{s \in S} \;  \left[\, x_s \, a_{s,x}(T) + \log( g(a_{s,x}(T) )\, \right].
\]

\subsection{Computation of the MPLE }\label{computeMPLE}
Let $X$ be a random configuration with autologistic distribution $\pi_\theta$. For all $T \in \Theta$, define the \textit{mean log pseudo-likelihood} $ g_\theta(T) $ by 
\begin{equation}\label{gtheta}
g_\theta(T) = E_\theta (LPL(X,T)) = \sum_{s \in S} \sum_{x \in \mathcal{B}} \pi_\theta(x) \, \left[ \, x_s \, a_{s,x}(T) + \log( g(a_{s,x}(T) ) \, \right].
\end{equation}
The theoretical principle of the MPLE algorithm is to seek a vector estimate $T$ of $\theta$ which maximizes in $T$ the \textit{mean log pseudo-likelihood}. This approach relies on the strict concavity of $g_\theta (T)$ as a function of $T$ (see Proposition \ref{propconcave} below). 

Consider $n$ independent observed configurations $\{ X^{1}, \ldots, X^{n}\}$, generated by the same unknown autologistic distribution $\pi_\theta$. Due to the law of large numbers one can approximate the unknown function $g_\theta(T) = E_\theta ( LPL(X,T) )$ by the  \textit{empirical log pseudo-loglikelihood} 
\begin{equation}\label{MPLEobj}
\hat {g}(T,n) = \frac{1}{n} \sum_{j=1}^n LPL( X^{j}, T ).
\end{equation}
Let $\partial_T LPL(x, T)$ and $Hess_T LPL(x, T)$ denote the gradient and Hessian matrices of $LPL(x, T)$ with respect to $T$, then the gradient and Hessian matrix of $\hat {g}(T,n) $  with respect to $T$ are 
\begin{equation}\label{hatgradG}
\hat{G}(T,n)= \frac{1}{n} \sum_{j=1}^n \partial_T LPL( X^{j}, T ),
\end{equation}
and
\begin{equation}\label{hathessH}
\hat{H}(T,n)= \frac{1}{n} \sum_{j=1}^n Hess_T LPL( X^{j}, T ).
\end{equation}

The Maximum Pseudo-Likelihood Estimator $\hat{\theta}(n) $ of $\theta$ is defined as a vector which minimizes $ \hat {g}(T,n) $ in $T$, and hence verifies the non linear vector equation
$$
\hat{G}(\hat{\theta}(n), n)= 0.
$$
The existence and fast computability of $\hat{\theta}(n)$ is due to the following proposition.

\begin{proposition}\label{propconcave}
 For each $\theta \in \Theta$, the mean log pseudo-likelihood $g_\theta (T)$ is  a strictly concave function of $T \in \Theta$, and reaches its maximum in $T$ at the unique point $T= \theta$. Moreover, the empirical pseudo-likelihood $ \hat {g}(T,n) $ is also concave in $T \in \Theta$, and becomes almost surely strictly concave as $n \to \infty$. 
\end{proposition}

\textbf{Proof:} The proof is given in Appendix \ref{MPLEexist} of the Supplementary Materials.

Due to the concavity of $ \hat {g}(T,n) $,  we implemented a standard gradient descent to generate a sequence $T(j)$ converging to $\hat{\theta}(n)$ as $j$ increases
\[
T(j+1)= T(j) + \epsilon\;\hat{G}(\,T(j),n\,).
\]
One stops iterating when $ || \, \hat{G}(\,T(j),n\,) \, || $ becomes inferior to a user-chosen small threshold. 

For each one of our  benchmark discrimination studies, autologistic modeling by MPLE was used intensively to parametrize  roughly around 2,000 to 5,000 Gibbs models of dimension less than 30, in order to explore enough  potentially discriminating combinations  of sites and cliques of order 2, automatically selected among the large list of reference biomarkers. Each MPLE modeling was implemented by gradient descent involving 200 iteration steps, with a step size $\epsilon=0.05$. On a 1.3 GHz MacOS PC, the average computing time per Gibbs model was about 1 second for colorectal cancer stage groups of mass spectra, which  all had small  size $\leq 80$; this CPU time increased to about 8 seconds for groups of size $\sim 100$, such as our ovarian cancer groups, and reached about 15 seconds for groups  of size $\sim 200$, such as the full colorectal cancer group.

\subsection{Asymptotic Normality of MPLE}\label{estimerror}
Most early results on asymptotic normality of the MPLE have focused on Gibbs random fields on infinite lattices under Dobrushin unicity conditions (see for instance \cite{Jensen1994}). For Gibbs random fields on the finite set $\mathcal{B}$ of binary vectors of fixed length $L$, the Dobrushin conditions become irrelevant, so that many publications consider  asymptotic consistency and normality of MPLEs as valid for finite configurations spaces, but without refering to explicit proofs. We state a precise asymptotic result, proved in the  mathematical Appendix \ref{asymnorm} of the Supplementary Materials.

\begin{proposition}\label{propasym}
 For any autologistic distribution $\pi_\theta$ on the set of binary configurations $\mathcal{B}$, the MPLE estimators $\hat{\theta}(n)$ of  $\theta$ are asymptotically consistent as the number of observations $ n \to \infty$. The normalized vectors of estimation errors $\sqrt{n} \; (\hat{\theta}_n - \theta)$ are asymptotically Gaussian with mean zero and a covariance matrix $\Gamma(\theta)$ computable as indicated in the Appendix \ref{asymnorm} of the Supplementary Materials. 
\end{proposition}

\textbf{Proof:} The proof is given in Appendix \ref{asymnorm} of the Supplementary Materials.

\subsection{Elimination of non-significant parameters}\label{elimination} 
The preceding asymptotic normality  result provides a tool to decide if some estimated parameter coordinates should be replaced by $0$. Indeed, the diagonal elements $\Gamma_s$ and  $\Gamma_{s,t}$ of the covariance matrix $\Gamma(\theta)$ can easily be approximated from the data (see Appendix \ref{asymnorm}).  The standard deviations $\hat{\theta}_s(n) - \theta_s$  and $\hat{\theta}_{s, t}(n) - \theta_{s, t}$ are then $\sigma_s(n) \sim \sqrt{\Gamma_s }/ \sqrt{n}$ and $\sigma_{s,t}(n) \sim \sqrt{\Gamma_{s,t}} / \sqrt{n}$, which yields explicit Gaussian 90\% confidence intervals for $\theta_s$  and $\theta_{s, t}$. Whenever one of these 90\% confidence intervals contains $0$,  we constrain the corresponding coordinate of $\hat\theta$  to be $0$. The autologistic model is then re-estimated  by MPLE, but the gradient descent implementing MPLE computation now takes into account the complementary constraints just introduced on $\theta$. This procedure is iterated until all estimated parameters are significantly different from $0$. 

In our benchmark studies, the sizes $n=54 , 80 , 74 , 30$ of the  four colorectal cancer data sets ADE, ECR, LCR CTR were rather small. So in the autologistic modeling of these 4 cancer datasets the relative  accuracies $ \sqrt{\Gamma_s }/ (\sqrt{n}\hat{\theta}_s) $ of several  estimated coordinates $\hat{\theta}_s$  ranged  between 30\% and 50\% (see for instance the accuracy results displayed in Table \ref{tab:coeff} below). It would clearly have been desirable to double the sizes of our colorectal cancer data sets, but the mass spectra acquisition phase  had  already been fully terminated when the present statistical study began. Our four ovarian cancer groups have larger sample sizes $n=116 , 100,  91, 162$ which naturally led to more accurate modeling. Each one of our  benchmark studies involved the estimation of at least two thousand autologistic models, and a quick analysis of all the associated  estimation accuracy results showed that  ``ideal" datasets sizes  $n \sim 300$ would be amply sufficient for very accurate modeling. 

Consider two groups $G^+$ and $G^-$ of binarized mass spectra, with resp. sizes $n^+$ and $n^-$. For automatic discrimination between these two groups, we will model both of them  by autologistic distributions $\pi^+$ and $\pi^-$. Modeling accuracy  will not be the main criterion , since the goal is then to identify \textit{short} biomarker signatures enabling  high performance discrimination. The well known  parameter parsimony principle is here quite relevant, and suggests to strongly restrict  the total number $m$ of nonzero coordinates in the joint parametrization of $\pi^+$ and $\pi^-$. Ideally one should have $m << n^+ + n^-$. In our benchmark discrimination tasks, our optimized signature selection procedure  typically  yielded $5 \leq m \leq 30$ and $ m \leq (n^+ + n^-)/5 $.

\subsection{Quality of Fit: Empirical Estimation}\label{QF}
After fitting an autologistic distribution $\nu = \pi_{\hat{\theta}}$  to  a data set $G$ of $n$ observed binary vectors, the \textit{quality of fit} of $\nu$  to the data can be  evaluated as follows.

First compute the empirical log-likelihood $LL $ of the data set $G$ under the probability $\nu$. To approximate the distribution $\lambda$  of $LL$ under $\nu$, we implement a classical Gibbs sampler (\cite{Bremaud1999}) based on the Gibbs distribution $\nu$ to simulate $1000$ virtual data sets $\mathcal{G}_1, \mathcal{G}_2, \ldots, \mathcal{G}_{1000}$, where each $\mathcal{G}_j$ contains $n$ random binary vectors having  the same  probability distribution $\nu$. Then compute the log likelihood $LL_j$ of each virtual data set $\mathcal{G}_j$ under the probability $\nu$. After re-ordering the list of 1000 log-likelihood values $LL_1,L L_2, \ldots, LL_{1000}$, the rank of $LL$ becomes $1 \leq r(LL) \leq 1000$. 

For the log likelihood distribution $\lambda$ the number   $Q= r(LL)/1000$ is the (random)  quantile corresponding to the observed $LL$. Under the true but unknown probability $\pi_\theta$ and for $n$ large, the quantity $\sqrt{n}\; (Q - 50\%)$ is approximately Gaussian with mean zero, and hence can easily be used to quantify the goodness of fit of $\pi_{\hat{\theta}}$ to the data set $G$. Namely, a good quality of fit between  $\pi_{\hat{\theta}}$ and $G$ should correspond to a $Q$ value close to 50\%.

\begin{figure}[!h]
\centering\includegraphics[width=\textwidth]{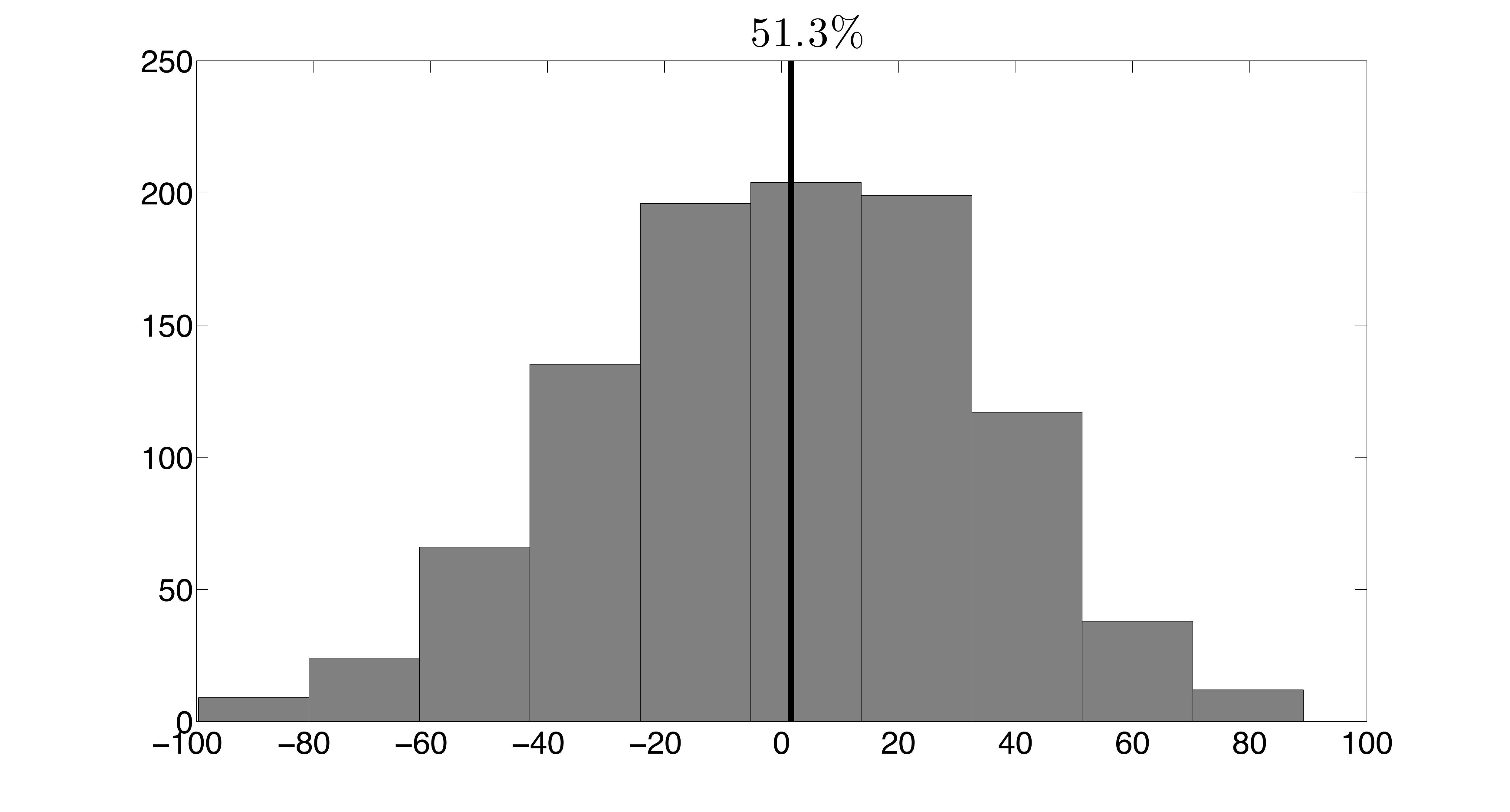}
\caption{This figure displays the quality of fit of an autologistic model  $\nu = \pi_{\hat{\theta}}$ fitted to a data set of 74 mass spectra, namely the colorectal cancer group LCR. The 18 biomarkers involved in this model were selected (see details further on) to  optimize the discrimination ECR vs LCR. The log likelihood histogram displayed here is based on  1000 virtual log likelihood values resp. computed on 1000 simulated random samples of 80 binary vectors of length 18, simulated from $\nu $ by Gibbs sampler. The bold vertical black line correspond to the true log likelihood value computed on the actual 74 binarized spectra in LCR.}\label{fig:histMPLE}
\end{figure}

In our benchmark studies, the autologistic distributions estimated by MPLE generally have very good quality of fit to real mass spectra datasets. For example, in  our  colorectal cancer data, to discriminate the patients groups $G=LCR$ from $ECR$, the algorithms described below identified the  ``best" autologistic model $\nu$ for the group $G$ (which contained $n= 74$ mass spectra). The model $\nu$ involved only 18 optimally selected biomarkers and the quantile $Q = r(LL)/1000$ had the value $51.3\%$,   which indicates a very good quality of fit. Figure \ref{fig:histMPLE} displays the histogram of  1000 virtual log likelihood values $LL_j$ generated by Gibbs sampler   based on $\nu$,  the bold vertical black line has abscissa $Q=51.3\%$.

\section{Optimal Discrimination between Autologistic Models}\label{optidis}
On a set $\mathcal{B}$ of binary vectors with length $L$, let $\pi^+= \pi_{\theta^+}$ and 
$\pi^- = \pi_{\theta^-}$ be two autologistic distributions, parameterized by $\theta^+, \theta^- \in \Theta$, and with partition functions $Z^+ = Z(\theta^+) $ and  $Z^- = Z(\theta^-)$. We are using the same set of cliques for $\pi^+$ and $\pi^- $(namely, $\pi^+$ and $\pi^- $ share the same $U(x)$), which is not a restriction since the coordinates of $\theta^+$ and $ \theta^- $ are allowed to take the value zero. 

Call \textit{decision rule} any real valued function $g$ defined on $\mathcal{B}$. Each such $g$ classifies  random  observation $x \in \mathcal{B}$ as ``generated by $\pi^+$" if $g(x) > 0$, and  as ``generated by $\pi^-$" if  $g(x) < 0$. The \textit{performance} of any decision rule $g$ is quantified by the two probabilities $p^+(g)$ and $p^-(g)$ of correct decisions when $g$ is confronted to random configurations respectively generated either only by $\pi^+$ or only by $\pi^-$, so that 
\[ 
p^+(g) = \pi^+[ \, x  \mid g(x) >0 \, ] \ \ \textit{and} \ \ 
p^- (g)= \pi^- [ \, x  \mid  g(x) < 0 \, ].
\]
Optimal decision rules are  characterized as follows.

\textbf{Optimal  Decision Rules:} Fix any weight coefficient $0 < \alpha <1$. Among all decision rules $g$, there is an \textit{optimal decision rule} $f$ which maximizes $( \alpha \, p^+(g) + (1 - \alpha) \, p^-(g) )$. More precisely, there is then a number $\gamma$ determined by $\alpha$, such that for all $x \in \mathcal{B}$, the optimal $f$ verifies 
\[f(x) =\, < u, \, U(x) > +\, a,\quad \textit{with}\quad u=  \theta^-- \theta^+ \quad \textit{and}\quad a = \log( Z^- / Z^+) - \gamma. \]
When $\alpha = 1/2$, the \textit{performance criterion} to be maximized becomes $ PERF(g)= ( p^+(g) + p^-(g) ) / 2$ and the number $\gamma$ is actually zero. 

\textbf{Proof of this characterization:} The relation $f(x) > 0$ is equivalent to $D(x) > e^{\gamma}$ where the likelihood function $D(x)$ is the density function $\frac{d \pi^+}{d \pi^-}(x) $.  The announced result can then be derived from the Neyman-Pearson lemma, essentially as was done  in Proposition 5.2 of \cite{Azencott2014}. 
When $\alpha = 1/2$, the symmetry   between $\pi^+$ and $\pi^- $ immediately shows that $\gamma=0$. 

\section{Discrimination errors due to estimation errors on model parameters}
\subsection{Estimated optimal separator } 
In  concrete  discrimination tasks, the autologistic models $\pi^+$ and $\pi^-$ are not known but derived from independent observed configurations $X^1, \ldots, X^n$ and $Y^1, \ldots, Y^n$, which we consider as  separately  generated by $\pi^+$ and $\pi^-$. The unknown vectors of parameters $\theta^+$ and $\theta^-$ are  then naturally replaced by their MPLE estimators $\hat{\theta}_n^+$ and $\hat{\theta}_n^-$. In our benchmark applications below, we systematically maximize  the simplest \textit{performance criterion PERF(f)} $= ( p^+(f) + p^-(f) ) / 2$, so that due to the ``Optimal  Decision Rules'' in Section \ref{optidis}, the optimal separator $f$ is then given by
\begin{equation}\label{f(x)} 
f(x) = < u, \, U(x) > + a,\quad \textit{with}\quad u=  \theta^-- \theta^+ \quad \textit{and}\quad a = \log( Z^-/Z^+). 
\end{equation}

After replacing the parameters of $( \pi^+ , \pi^- )$ by their  MPLE estimates, the unknown  models  and their partition functions are replaced by their estimates $( \hat{\pi}^+ , \hat{\pi}^- )$ and $( \hat{Z}^+ , \hat{Z}^- )$. The optimal separator $f$ is estimated by $\hat{f}$ where 
\begin{equation} \label{hatf(x)}
\hat{f}(x) = < \hat{u},\, U(x) > +\, \hat{a} \quad \textit{with}\quad \hat{u}=   \hat{\theta}_n^- - \theta_n^+,\quad \textit{and} \quad \hat{a} =  \log( \hat{Z}^- / \hat{Z}^+ ). 
\end{equation} 
 When the classification of observed binary vectors $x$  is  based on the sign of $\hat{f}(x)$ instead of sign($f(x)$), the  performance quantifiers $p^+(f)$ and $p^-(f)$ are  replaced by $\hat{p}^+$ and $\hat{p}^-$ given by 
\[ 
\hat{p}^+ = \pi^+ \, [ \, x  \mid \hat{f}(x) >0 \, ] \quad \textit{and} \quad 
\hat{p}^- = \pi^- \, [ \, x  \mid  \hat{f}(x) < 0 \, ].
\]

We are now going to estimate the errors $( \hat{p}^+ - p^+ )$ and $( \hat{p}^- - p^- )$.

\begin{proposition}
 In the preceding situation, as $n \to \infty$, the normalized error vector $\sqrt{n} ( \hat{u} - u )$ is asymptotically Gaussian with mean zero and covariance matrix $Cov = \Gamma(\theta^+) + \Gamma(\theta^-) $, where the matrices $\Gamma(\theta)$ have been computed in Proposition \ref{propasym}. Moreover, $\sqrt{n} \, ( \hat{a} - a )$ is also asymptotically Gaussian with mean zero and asymptotic variance $var(\theta^+, \theta^-)$ verifying 
\begin{equation}\label{var}
\var(\theta^+, \theta^-) \leq 
\frac{L(1+L)}{2} \times trace \, \left[ \, \Gamma(\theta^+) + \, \Gamma(\theta^-) \right].
\end{equation}
\end{proposition}

\textbf{Proof:} The proof is given in Appendix \ref{estisep} of the Supplementary Materials.

We now compare the discriminating powers of decision regions defined by two affine separators.

\begin{proposition}\label{properror}
 On the set $\mathcal{B}$ of binary vectors $x$ with length $L$, consider the two  separators 
\[
f(x)= < u, U(x) > + a \; , \quad \phi(x)= < \eta, U(x) > + \alpha 
\]
parametrized by $u, \eta \in {\Theta}$ and  $a, \alpha \in R$.
Fix any autologistic  distribution $\pi_\theta$ on $\mathcal{B}$. Let $p(f)$ and $p(\phi)$ be the two probabilities 
\[
p(f) = \pi_\theta [ \, x \mid f(x) >0 \, ] \quad\textit{and}\quad
p(\phi) = \pi_\theta [ \, x \mid \phi(x) >0 \, ].
\]
Recall that $k = L(1+L)/2$. Fix $u,a, \theta$ and assume that $0 < p(f) < 1$. For any arbitrary small percentage $0\% \leq \gamma < 100\%$, there is a strictly positive number $q$ depending only on $\theta, u, a, \gamma$ such that for any pair $(\eta, \alpha)$ verifying
\begin{equation} \label{cond}
|| \eta -u || < \frac{q}{2\sqrt{k}} \quad \textit{and} \quad | \, \alpha - a \, | < q/2,
\end{equation} 
one must then have $ | \, p(f) - p(\phi) \, | \leq \gamma$.
\end{proposition}

\textbf{Proof:} The proof is given in Appendix \ref{estisep} of the Supplementary Materials.

\subsection{Control of discrimination errors due to separator estimation}
\begin{proposition}  Fix two autologistic distributions $\pi^+$and $\pi^- $ parameterized by $\theta^+, \theta^- \in \Theta$. Let $f(x) = < u,\, U(x) > +\, a$ be the optimal separator between $\pi^+$ and $\pi^-$, given by equation \eqref{f(x)}. From two random samples of $n$ configurations resp. generated by $\pi^+$ and $\pi^-$, one computes by equation \eqref{hatf(x)} the estimator $\hat{f}(x) = < \hat{u},\, U(x) > +\, \hat{a}$ of $f(x)$. 

Let $ p^+(f), p^-(f) $ and $ p^+(\hat{f}), p^-(\hat{f}) $ be the probabilities of correct  discrimination between $\pi^+$ and $\pi^-$ resp. achievable by the separators $f$ and $\hat{f}$. Assume that $0 < p^+(f) < 1$ and $0 < p^-(f) < 1$. 

Let $\gamma$ and $\kappa$ be two arbitrary small numbers verifying $0 \leq \gamma < 1$
and $0 < \kappa < 1$. Then one can find $N$ such that for $n > N$ one has 
\begin{equation} \label{result}
P (\; | \, p^+(\hat{f})  - p^+ \, | \leq \gamma \;\; \textit{and} \;\;
| \, p^-(\hat{f}) - p^- \, | \leq \gamma \; ) \; \geq \; 1 - \kappa 
\end{equation}
within the following proof.
\end{proposition} 

\textbf{Proof:} The proof and a practical estimate of $N$ are given in the Appendix \ref{ctrerror} of the Supplementary Materials.

\section{ Markov Random Field (MRF)  Discrimination}\label{MRFdiscrim}

\subsection{Biomarker Selection to Discriminate Between two Groups of Mass Spectra}\label{feature}

Concrete datasets of mass spectra acquired from cancer patients usually involve several distinct patient groups, but are often of moderate sizes inferior to 100 spectra, as in our benchmark studies below. For practical discrimination between several patient groups, our pre-processing of raw mass spectra is done simultaneously for the data of all these groups. This generates a list of reference biomarkers $B_s$ with $s \in S$, where $S$ typically has large size $L$. For instance further below, in our colorectal cancer study, we have $L \sim 800$, and for the benchmark ovarian cancer data, pre-processing by other teams had yielded $L \sim 15,000$.

When fitting autologistic models $\pi^+$, $\pi^-$ to  two training sets  $BinG^+$, $BinG^-$ of binarized mass spectra, one wants to achieve statistically \textit{robust} fitting, as well as high discriminating power between $\pi^+$ and  $\pi^-$. Among the set of reference biomarkers $B_s, s \in S$, one must hence select a subset  of small cardinal $d$, and for each selected $B_s$, the presence or absence of $B_s$ within any $x$ in $BinG^+$ or $BinG^-$ should provide strongly discriminating information about the correct classification of $x$. This leads us to the following \textit{feature selection} algorithm.

Given two training sets $BinG^+$ and $BinG^-$ of binarized spectra, both included in the set of binary vectors $\mathcal{B}(L)$,  we compute for each  $s \in S$ the resp. frequencies $m^+(s)$ and $m^-(s)$ of the event $x_s = 1$ for $x \in BinG^+$ and for $x \in BinG^-$.  To  ensure a significant presence of  each selected biomarker $B_s$ in at least one of the two groups, we fix a minimal frequency threshold $thr$, taking account of the number $n$ of training data, and we select in $S$ the subset $\hat{S}$ of all sites $s$ for which $\; \min \{ m^+(s),  m^-(s) \}\geq thr \;$. For our benchmark studies  where $30 \leq n \leq 162$ was rather small, we systematically used   $thr = 20\%$.

As in  \cite{Kong2014}, we quantify the  \textit{Discriminating Power} of each site $s$ in $\hat{S}$ by the ratio $DP(s) = m^+(s)/ m^-(s)$. This empirical definition is easily justified when one discriminates between two autologistic distributions where all pairs of sites are weakly correlated. Highly discriminating biomarkers $B_s$ tend to have either very high or  very low $DP(s)$  values. Since we want to discover highly discriminating signatures involving only a small number $d$ of biomarkers,  we  fix a moderate number $H$ (approximately equal to $1\%$ of cardinal$(\hat{S})$ in all our benchmark applications), and we  focus only on the $H$ sites $s \in \hat{S}$ with the highest  $DP(s)$ and the $H$ sites with the lowest  $DP(s)$.

Fix  any two positive integers $d^+ \leq H$ and $d^- \leq H$, so that   $d << L$. The choice of $d^+, d^-$ will be optimized further on. Within the biomarkers $B_s$ with $s \in \hat{S}$, we select as ``$G^+$ biomarkers" the $d^+$ biomarkers $B_s$ with the highest $DP(s)$, and as ``$G^-$ biomarkers" the $d^-$ biomarkers $B_s$ with the lowest $DP(s)$. 

The union $ S(d)\subset \hat{S}$ of the two sets of $d^+$ and $d^-$ sites just selected has cardinal $d = d^+ + d^- << b$. Each binary vector $x$ in $BinG^+$ or $BinG^-$ is then systematically \textit{restricted} to the $d$ sites of $S(d)$. This restriction generates two sets of binary vectors in $\mathcal{B}(d)$ which we will denote by $BG^+$ and $BG^-$.

\subsection{Clique Discovery for two Autologistic Models}\label{clique}
Given the data set $BG^+$ of $n$ binary vectors, and any pair of sites $s,t$, one can easily compute   the empirical joint frequencies of the four events $\{ X_s = i, X_t = j \}$, where $i$ and $j$ have binary values $0$ or $1$. We use this $2 \times 2$ contingency table to  quantify the stochastic dependency between $X_s$ and $X_t$ by a classical $\chi^2$-statistic with one degree of freedom, denoted $\chi^2(s,t)$. At the 95\% significance level,  $X_s$ and $X_t$ are thus considered ``dependent"  iff $\chi^2(s,t) > 3.84$, and the pair   $s, t$. will then be retained as a potential clique.

The feature selection procedure used to generate  $BG^+$ and $BG^-$ (see Section \ref{feature}) implies that whenever $B_s$ is a $G^-$ biomarker, the frequency $m^+(s)$ of the event $X_s =1$ within the dataset $BG^+$ will typically be quite small. For all our benchmark studies, we had indeed $m^+(s) < 5\%$ for all $G^-$ biomarkers $B_s$. Practically, for mass spectra acquired from cancer patients, datasets sizes $n$ are moderate, of the order of  100. So when $m^+(s) < 5\%$, it is  hopeless to obtain reliable estimates of the dependency statistics $\chi^2(s,t)$ for any $t$. 

To achieve a reasonably robust fitting of an autologistic model $\pi^+$ to the dataset $BG^+$, 
the set $\mathcal{C}^+$ of cliques of order 2 for $\pi^+$ will hence be restricted to include only  pairs $\{s, t\}$  such that $\chi^2(s,t) > 3.84$ and both $B_s$ and $B_t$  are $G^+$ biomarkers. Call  $\mathcal{PTC}^+$ the set of all such potential cliques. The cardinal  $c^+$ of $\mathcal{C}^+$ will successively  be fixed at any value inferior to $card(\mathcal{PTC}^+)$, and will be optimized further on. Once we fix $c^+$, we  retain in $\mathcal{C}^+$ precisely the $c^+$ cliques $\{s, t\}$ belonging to  $\mathcal{PTC}^+$ and having the $c^+$ highest $\chi^2(s,t)$ values. 

Similarly, let $\mathcal{PTC}^-$ be the set of pairs $B_s, B_t$  of  $G^-$ biomarkers such that $\chi^2(s,t) > 3.84$ ,  with $\chi^2(s,t)$ evaluated on $BG^-$. To fit $\pi^-$ to $BG^-$, we tentatively fix any $c^-\leq card(\mathcal{PTC}^-)$, to be optimized later on, and the set of cliques $\mathcal{C}^-$ for $\pi^-$ are the $c^-$ cliques $\{s, t\}$ with highest $\chi^2(s,t)$ in $\mathcal{PTC}^-$.

\subsection{Numerical Autologistic Model Fitting}\label{numautomodel}
Call $\mathcal{D}(H)$ the set  of all  quadruplets of integers $dim= (d^+, c^+; d^-, c^-)$ such that 
\begin{equation}\label{D(H)}
1 \leq d^+ \leq H \, , \;\; 1 \leq d^- \leq H \,, \;\; 0 \leq c^+ \leq card(\mathcal{PTC}^+)\, , \;\; 
0 \leq c^- \leq card(\mathcal{PTC}^-)\,,
\end{equation}
where the moderate integer $H$ was selected in Section \ref{feature}. The set $\mathcal{D}(H)$ has size $\sim O(H^2)$  which  increases quickly with $H$.

For each quadruplet $dim= (d^+, c^+; d^-, c^-)$ in $\mathcal{D}(H)$, the procedure described in Section \ref{feature} selects a number $d^+ $ of $G^+$ sites and a number $d^- $ of $G^-$ sites which determine a set of sites $S(d)$ of cardinal $d= d^+ + d^-$. The procedure given in Section \ref{clique} then selects a set $\mathcal{C}^+$ of $c^+$ cliques among pairs of $G^+$ biomarkers, and a set $\mathcal{C}^-$ of $c^-$ cliques among pairs of $G^-$ biomarkers. 

On the set $\mathcal{B}(d)$ of binary vectors  indexed by $S(d)$, we can then define  as follows two autologistic models $\pi^+$ and $\pi^-$ parameterized by the unknown parameter vectors $\theta^+$ and $\theta^-$ 
\[
\pi^+=\pi_{\theta^+}(x) =\frac{1}{Z(\theta^+)} e^{- < \theta^+, U(x) >} \quad \textit{and}\quad
\pi^- = \pi_{\theta^-}( x) = \frac{1}{Z(\theta^-)} e^{- < \theta^-, U(x) >}, 
\]

with
\[
<\theta^+, U(x)> =\sum_{s\in S(d)} \theta^+_s x_s  + \sum_{\{s, t\}\in \mathcal{C}^+}\theta^+_{s, t} x_s  x_t, \quad \textit{and}\quad<\theta^-, U(x)> =\sum_{s\in S(d)}\theta^-_s x_s + \sum_{\{s, t\}\in \mathcal{C}^-} \theta^-_{s, t} x_s x_t.
\]
To fit the  models $\pi^+ = \pi_{\theta^+}$ and $\pi^- = \pi_{\theta^-}$ to the two data sets $BG^+ \subset \mathcal{B}(d) $ and $BG^- \subset \mathcal{B}(d) $, we estimate $\theta^+$ and $\theta^-$ by the MPLE technique outlined in Section \ref{MPLE}. Two separate gradient descent algorithms implement the computation of  estimates $\hat{\theta}^+$ and 
$\hat{\theta}^-$. As explained in Sections \ref{estimerror} and \ref{elimination}, after computation of the 90\% error margins on these estimated parameters, the coordinates of $\hat{\theta}^+$ and $\hat{\theta}^-$ which are not significantly different from $0$ are then forced to be zero.

This autologistic model fitting procedure is repeated for each quadruplet $dim \in \mathcal{D}(H)$. This generates a number $N\sim O(H^2)$ pairs of  autologistic models 
$\pi^+$ and $\pi^- $, all derived from the same original  mass spectra  data sets $G^+$ and 
$G^-$.
\subsection{Numerical Estimation of Optimal Separator}\label{sysalgo}

Fix any quadruplet $dim= (d^+, c^+; d^-, c^-)$ in $\mathcal{D}(H)$. Let $\pi^+ $ and $\pi^-$ be the two autologistic models associated to $dim$ and resp. fitted to the data sets $BG^+$ and $BG^-$. 

Formula \eqref{f(x)} for the optimal decision rule $f(x)$ discriminating between $\pi^+ $ and $\pi^-$ involves the term $\log(Z^- / Z^+)$. But computing   the partition function $Z(\theta)$  of any autologistic $\pi_\theta$ by   summing the $v(x) = e^{<\theta, U(x)>}$ over all $x \in \mathcal{B}(d)$   quickly becomes a heavy numerical task for $d > 16 $. A known faster approach,  justified by the law of large numbers, is to estimate  $Z(\theta)/ 2^d $ by the  average   of $v(W_1), \ldots, v(W_N)$ where $N \sim 10,000$ and the $W_j$ are independent random binary vectors generated by the \textit{uniform} distribution on $\mathcal{B}(d)$.  To avoid the errors involved in estimating  $\log(Z^- / Z^+)$, and to partially compensate for  parameters estimation errors, we have preferred to implement the following algorithm.

For each binary vector $x$ in $\mathcal{B}(d)$, define the planar point $W(x)$ in $ R^2$ by 
\begin{equation}\label{W}
W(x) = ( w^+(x), w^-(x) ) \quad\textit{with}\quad w^+(x) = <\hat{\theta}^+, U(x)> \quad\textit{and}\quad w^-(x) = <\hat{\theta}^-, U(x)>.
\end{equation} The non linear  function $x \to W(x)$ thus transforms the subsets $BG^+$ and $BG^-$ of $\mathcal{B}(d)$ into two sets $PLG^+ \subset R^2$ and $PLG^- \subset R^2$ of planar points labeled by $+1$ when $x \in BG^+$ and $-1$ when $x \in BG^+$. The optimal  separator $ f(x) $ defined by \eqref{f(x)} is equivalent to an affine function $\mathcal{A}$ of $W(x) \in R^2$ given by 
\begin{equation}\label{linearNPf}
\mathcal{A}. W(x) = f(x) =   w^-(x) - w^+(x) + Constant.
\end{equation}
Therefore, to generate a robust estimate $\hat{f}$ of  the optimal separator $f$, we can directly search for an affine function  $\mathcal{A}$ defined on the plane $R^2$, and which separates the two finite sets $PLG^+$ and $PLG^-$ with a small number of  errors. 

To compensate for the errors due to the estimations of $\theta^+$ and $\theta^-$, which generate errors on the planar point  $W(x)$, we introduce  an error correcting coefficient  $\beta$ in  the affine  separator $\mathcal{A}$ as follows, 
\begin{equation}\label{corrected.hat.f}
\mathcal{A}. W(x) = \hat{f}(x) =  \beta w^-(x) - w^+(x) + Constant,
\end{equation}
where $\beta$ can be slightly different from 1. Given the two planar sets $PLG^+$ and $PLG^-$ the ``best" affine separator between $PLG^+$ and $PLG^-$ can be quickly computed by many classical affine discrimination algorithm, such as Support Vector Machines with linear kernels (\cite{Fan2008}). In each discrimination task $G^+$ vs $G^-$, the computation of the  best separator $\hat{f}$ between $\pi^+ $ and $\pi^-$ has to be repeated  for each quadruplet  $dim$ in $\mathcal{D}(H)$.  So to gain in CPU time, we estimate $\mathcal{A}$ by linear regression of an indicator matrix (\cite{Hastie2009}).

\subsection{ Performance Evaluation  for Autologistic  Separator}\label{perfeval} 
For each fixed quadruplet  $dim$ in $\mathcal{D}(H)$, we need to evaluate the probabilities $p^+(dim)$ and $p^-(dim)$ of successful discrimination  between $G^+$ and  $G^-$ based on the ``ideal" but unknown decision rule $f$. These probabilities  can be evaluated by classical leave-one-out cross validation (\cite{Geisser1993}). 

Namely, at each cross validation round, one single binarized spectrum $x(M)$ is temporarily eliminated from $BG^+ \cup BG^-$, and this modified dataset is used to generate (as just outlined above) an estimated separator $F= \hat{f}$ of $f$. One checks then the sign of $F(x(M))$, to record whether the ``left out" $x(M)$ is correctly classified by $F$ or not. This procedure is repeated until every $x(M)$ in the dataset has been left out once. The \textit{leave-one-out estimates} $\hat{p}^+$, $\hat{p}^-$ of $p^+(dim)$, 
$p^-(dim)$ are then the respective percentages of correct classifications  of $x(M) \in BG^+$ and of $x(M) \in BG^-$. The \textit{empirical performance} $PERF(dim)$ is then evaluated by $(\hat{p}^+ + \hat{p}^-)/2$.

The best choice $dim_{opt}$ for the quadruplet $dim = (d^+, c^+; d^-, c^-)$ is then ideally determined by maximizing the empirical performance $PERF(dim)$ over all $dim$ in $\mathcal{D}(H)$. However, since the cardinal of $\mathcal{D}(H)$ ranges between 2,000 and 5,000 in all our benchmark studies,  the full leave-one-out computation is too costly to be used for the computation of \textit{all} the $PERF(dim)$ values. Fortunately, it is unnecessary to do so in numerical implementations, as will be seen  in Section \ref{acceleration} below.

\subsection{Optimal Signature and associated Scores}\label{signatures}
For the discrimination task $G^+$ vs $G^-$, once the best quadruplet $dim_{opt} = (d^+, c^+; d^-, c^-)$ has been determined by performance maximization,  one has immediate access to  the corresponding best  autologistic models  $\pi^+_{opt}$ and $\pi^-_{opt}$, based on the selection of $d= d^+ + d^-$ specific reference biomarkers $ B_{s(1)}, \ldots , B_{s(d)}$, involving  numbers $d^+$ and $d^-$ of $G^+$ and  $G^-$ biomarkers, as well as $c^+$ pairs of  $G^+$ biomarkers and  $c^-$ pairs of $G^-$ biomarkers.

The list $ Sig = \{ B_{s(1)}, \ldots , B_{s(d)} \}$ then constitutes an explicit optimized \textit{signature} of length $d$ to discriminate between $G^+$ and $G^-$. 

Indeed let $\hat{f}_{opt}$ be the best  separator between $\pi^+_{opt}$ and $\pi^-_{opt}$,  as computed in \eqref{hatf(x)}.  For any mass spectrum $M$, let  $X(M)$ be the binarized vector associated to $M$, and let $x(M)$ be the restriction of this  binarized vector to the $d $ signature sites $ \left[ s(1), \ldots , s(d) \right]$. Classifying the  mass spectrum $M$ into $G^+$ or into $G^-$ requires only to know the sign of $F(M) = \hat{f}_{opt}(x(M))$. But the computation of $F(M))$ involves only checking  the presence or absence in $X(M)$ of each one of the biomarkers belonging to the signature $Sig$.

Moreover, classification of mass spectra can then be interpreted as follows. Let $\hat{\theta}^+$ and $\hat{\theta}^-$ be the vectors parametrizing $\pi^+_{opt}$ and $\pi^-_{opt}$. For any $j, k = 1 \ldots d$, define the \textit{scores} $SCO(j)$ and $SCO(j,k)$ 
\[
SCO(j) = \beta \hat{\theta}^-_{s(j)} - \hat{\theta}^+_{s(j)},\quad SCO(j,k) = \beta \hat{\theta}^-_{s(j), s(k)} - \hat{\theta}^+_{s(j), s(k)},
\]
where $\beta$ was estimated in \eqref{corrected.hat.f}. For any mass spectrum $M$, we add up the scores $SCO(j)$ of all biomarker $ B_{s(j)}$ which are present in $M$, as well as the scores $SCO(j,k)$ of all pairs of biomarker $ B_{s(j)}, B_{s(k)}$ which are jointly present in $M$. Adding to this sum the constant occuring in formula \eqref{corrected.hat.f} yields a \textit{total score} $TSCO(M)$, which is actually equal to $F(M)  = \hat{f}_{opt}(x(M))$, due to formulas \eqref{W} and \eqref{corrected.hat.f}. 

Hence $M$ is classified as belonging to $G^+$ or to $G^-$ according to the sign of its total score $TSCO(M)$.
\subsection{Accelerated Optimization of Discrimination Performances}\label{acceleration}
To estimate the best quadruplet $dim_{opt}$ by performance maximization in realistic computing time, we have developed and implemented the following accelerated optimization scheme. Note first that in all our benchmark studies, the 90\% error margins on $PERF(dim)$ are typically $\sim 0.08$.

For each quadruplet $dim$, once the autologistic models $\pi^+$, $\pi^-$, and their  separator $\hat{f}$ have been estimated,  one can instantaneously evaluate  the  empirical frequencies  of good decisions achieved  by $\hat{f}$ on the \textit{training sets} $G^+$ and $G^-$. The average $\widehat{PERF}(dim)$ of these two frequencies  provides a rough ``training" approximation  of $PERF(m)$. Once $\widehat{PERF}(dim)$ has been obtained for all $dim$, its maximization is immediate and provides first rough estimates $\hat{dim}_{opt}$ and  $\widehat{PERF}_{opt}$ for  the optimal quadruplet and the true best  performance. 

We can then compute $PERF(dim)$ by the costly  leave-one-out technique, but only on the quite smaller subset $\mathcal{SD}(H)$ of all quadruplets $dim$ which verify the constraint $\widehat{PERF}(dim) > \widehat{PERF}_{opt} -  0.08$. Maximizing $PERF(dim)$ for $dim$ in $\mathcal{SD}(H)$ gives us our final estimate for  $dim_{opt}$ and  the associated optimal performance. 

For each one of our six benchmark studies, the computing time for best signature discovery was thus reduced to a range of 3 to 9 hours on a 1.3 GHz MacOS PC.

\section{Kullback Distance Between Autologistic Models}\label{KLdistance}
Since the computation of performance $PERF(dim)$ is rather costly as seen above, we have also studied whether it was efficient to seek the best quadruplet $dim_{opt}$ by maximizing over all $dim \in \mathcal{D}(H)$ the  properly normalized Kullback-Leibler distance $norKL(dim) $ between  the two autologistic models $\pi^+$  and $\pi^-$ associated to each quadruplet $dim=(d^+, c^+; d^-, c^-)$. 

Recall that the  Kullback-Leibler (KL) divergence, introduced in \cite{Kullback1951}, defines a well known distance non-negative distance $KL(P, Q)$ between probability distributions $P$ and $Q$ on a finite set $\mathcal{B}$ (\cite{Dagan1999}, \cite{Bigi2003}, \cite{Bigi2003(2)}), given by 
\[
KL(P,Q)= \sum_{x \in \mathcal{B}} (P(x)-Q(x)) \, \log(\frac{P(x)}{Q(x)}).
\] 
and verifying $KL(P, Q)= 0$ iff $P=Q$. 

For autologistic models  $\pi^+ = \pi_{\theta^+}$ and $\pi^- = \pi_{\theta^-}$, we thus have 
\begin{equation}\label{dis2}
KL(\pi^+, \pi^-) = < \theta^- - \theta^+, \sum_{x \in \mathcal{B}} \; ( \pi^+(x) - \pi^-(x) ) U(x) >.
\end{equation}
Recall that $f(x) = < \theta^- - \theta^+, U(x) > + \log(Z^-/Z+) $ is the optimal separator between
$\pi^+ $ and $\pi^- $, we hence have 
\[
KL(\pi^+, \pi^-) = \sum_{x \in \mathcal{B}} f(x) \pi^+(x)  - \sum_{x \in \mathcal{B}} f(x) \pi^-(x).  
\]
When the optimal separators $f$ has high performance, $f(x)$ must be positive with high probability under $\pi^+$ and negative with high probability under $\pi^-$. In view of the last formula, we should then expect the distance $KL(\pi^+, \pi^-) $ to typically be high as well. 

This last  statement requires a proper normalization of KL distances, to take  into account the dimension $d= d^+ + d^-$ of the set of binary vectors $\mathcal{B}(d)$, since the autologistic probability distributions  $\pi^+$ and $\pi^-$ live on $\mathcal{B}(d)$. Indeed, a brief analysis of the case where pairs of sites are very weakly correlated suggests to normalize KL distances as follows 
\begin{equation}\label{norKL}
norKL(\pi^+, \pi^-) = KL(\pi^+, \pi^-) / \sqrt{d^+ +d^-},
\end{equation}
and for each dimension quadruplet $dim$, we will then denote 
$norKL(dim) =norKL(\pi^+, \pi^-)$.

Our MRF discrimination algorithm searches for ``well separated" pairs $\pi^+$ and $\pi^-$ by maximizing performance  $PERF(dim)$  over all dimensions quadruplets $dim$.  We conjecture that one can achieve roughly the same goal by  maximizing $norKL(dim)$ over all $dim$. Since $norKL(dim)$ is much easier to compute than $PERF(dim)$, the maximization of $norKL(dim)$ seems to be a rougher but faster approach to the discovery of the best quadruplet $dim_{opt}$.

As we have numerically checked in our six benchmark studies, this conjecture  is correct at the accuracy level  0.08 of all our performance evaluations. Indeed, our numerical results below show that normalized KL distances $norKL(dim)$ are   positively correlated to the actual discrimination performances $PERF(dim)$, and that this correlation  is stronger at  high values of $PERF(dim)$. 

\section{Application to Benchmark Cancer Datasets }

\subsection{Pre-Processing of Benchmark Mass Spectra}
For  the two ovarian cancer datasets, pre-processing had practically already been performed in \cite{Kong2014} and provided us with a fixed  list of  15,154 m/z ratios ranging from 0 to 20,000 and which thus defined our list of  15,154  reference biomarkers. 

For the colorectal cancer dataset,  we did implement the pre-processing of raw mass spectra as detailed in our joint paper \cite{Kong2014}, and this generated a list of 842 reference biomarkers positioned at the  842 m/z ratios $800 \times (1 + \rho)^j$ with  $\rho = 0.3\%$ and $j = 0, 1, 2, \ldots, 841$. 

As indicated in Section \ref{bincoding}, each pre-processed mass spectrum $M$ was then coded as a binary vector $x(M)$ of length 15,154 for ovarian cancer data, and of length 842 for colorectal cancer data.

\subsection{Benchmark Implementations of  MRF Discrimination}\label{MRFperf}
For each one of the 6 discrimination tasks outlined in Section \ref{data}, generically denoted $G^+$ vs $G^-$, we have systematically explored all the dimensions quadruplets $dim = ( d^+, c^+; d^-, c^- )$ belonging to $\mathcal{D}(H)$ where $H = 15$ for the three colorectal cancer stage discrimination tasks ADE vs ECR, ADE vs LCR, ECR vs LCR, and $H=10$ for the three cancer vs control discrimination tasks on both colorectal and ovarian cancer data. These deliberate restrictions forced two moderate a priori upper bounds for the numbers $d =d^+ + d^-$ of biomarkers explored  in signature discovery, namely signature lengths were thus kept inferior to $30$ for colorectal cancer data, and to $20$ for ovarian cancer data. Besides the associated reductions in  computing time, these bounds also forced the number of signature scores to remain  inferior to $n/5$ where $n$ was the total number of mass spectra in the training set $G^+ \cup G^-$.

For each quadruplet $dim \in \mathcal{D}(H)$, we have implemented our MRF discrimination algorithm to the pair $G^+, G^-$ of binarized mass spectra, as outlined in Section \ref{sysalgo}. This yielded two estimated autologistic models $\; \pi^+, \pi^- \;$ for $G^+, G^-$ and the associated best separator $\hat{f}$ between $\pi^+$ and $\pi^-$. The frequencies $p^+$ and $p^-$ of correct decisions achieved by $\hat{f}$ on the training sets $G^+$ and $G^-$ were then directly evaluated to derive a first rough estimate $\widehat{PERF}(dim) = ( p^+ + p^- )/2$ for the ``training performance" of the separator $\hat{f}$. After completing this computation of  $\widehat{PERF}(dim) $  for all $dim \in \mathcal{D}(H)$, the accelerated procedure  outlined in Section  \ref{acceleration} then enabled the computation of the optimal quadruplet $dim_{opt}= ( d^+, c^+; d^-, c^- ) $ maximizing $PERF(dim)$ and the associated separator $\hat{f}_{opt}$ discriminating between $G^+$ and $G^-$. 

As indicated in Section \ref{signatures}, this determined an optimized signature $Sig = \{ B_{s(1)}, \ldots , B_{s(d)} \}$ gathering  $d= d^+ + d^-$ specific reference biomarkers to discriminate $G^+$ vs $G^-$.  This computation also yielded numerical scores $SCO(i)$ and $SCO(i,j)$   associated to each signature biomarker $B_{s(i)}$ and to each retained clique $\{B_{s(i)}, B_{s(j)}\}$. As seen in Section \ref{signatures},  classifying a mass spectrum $M$ into $G^+$ or $G^-$ depends then only on the sign of its total score $TSCO(M)$, which  can be evaluated by checking the presence/absence in $M$ of the signature biomarkers.

\subsection{Benchmark Performances of MRF Discrimination } 

\begin{figure}[!h]
\centerline{\includegraphics[width=1.15\textwidth]{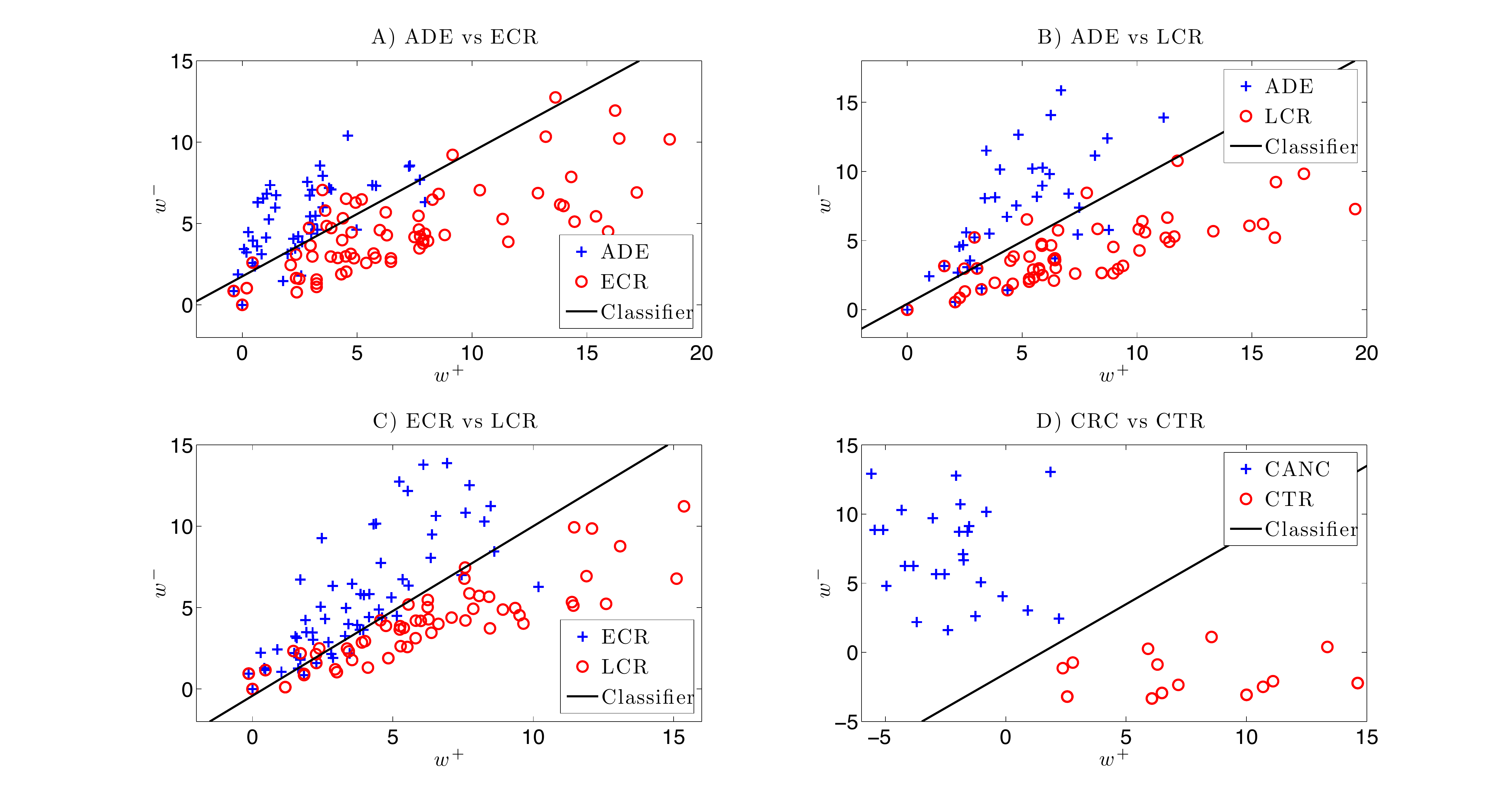}}
\caption{Colorectal Cancer Data: For each one of four discrimination tasks $G^+$ vs $G^-$ we display the planar representation of these two groups of mass spectra, derived from the optimized  autologistic models fitted to the data sets  $G^+$ and $G^-$. The associated  non linear recoding computes two planar coordinates $(w^+, w^-)$ for each mass spectrum. The best non linear separator between $G^+$ and $G^-$ can then be displayed as an affine function of $(w^+, w^-)$.}  \label{classifiers}
\end{figure}

As seen above, once the two optimized autologistic models $\pi^+, \pi^-$ have been estimated, the mass spectra datasets $G^+,G^-$ can be mapped into two sets $PLG^+$ and $PLG^-$ of planar points by the non linear map $M \to x = x(M) \to W(x)$, and the optimized classifier $\hat{f}_{opt}(x)$ is of the form $\mathcal{A}(W(x)$ where $\mathcal{A}$ is an affine function on $R^2$ which separates $PLG^+$ and $PLG^-$ with maximum margin. For each benchmark discrimination task on the colorectal cancer dataset, we have displayed $PLG^+, PLG^-$ and the associated best affine separator $\mathcal{A}$ in Figure \ref{classifiers}. 

Our MRF discrimination algorithm thus discovered 6 explicit optimized signatures, one for each benchmark discrimination task. We first note that these 6 signatures all had fairly short lengths $d= d^+ + d^-$, namely $d= 23, 17, 18, 7$ for the 4  colorectal cancer discrimination tasks, and  $d=12, 10$ for the two ovarian cancer discrimination tasks. These short signature lengths obviously are  a strong advantage for further biological interpretation of these 6 small families of key biomarkers, explicitly identified by their $m/z$ ratios.

The performances achieved on our 6 benchmark discrimination tasks by the optimal separators associated to these 6 signatures are displayed in Table \ref{tab:perf}. Taking into account the error margins on estimates of $p^+, p^-$,  our MRF discrimination algorithm reached performance levels $perf(MRF) =(p^+ + p^-)/2$ which were  \textit{essentially equivalent} to the best performances reported by previous publications, which used other discrimination algorithms.

\begin{table}[!h]
\tblcaption{Performance of optimized MRF Discrimination between $G^+$ and $G^-$ is evaluated by the frequencies $p^+$ and $p^-$ of correct  classifications within $G^+$ and $G^-$. For each discrimination task, we  display $p^+, p^-$ with their 90\% confidence intervals and the quadruplet $ ( d^+, c^+; d^-, c^- ) $ of dimension parameters  for the optimal autologistic separator between $G^+$ and $G^-$, which is based on a ``signature" involving  only $d= d^+ + d^-$ specific reference biomarkers.\label{tab:perf}}
{\tabcolsep=4.25pt
\begin{tabular}{@{}cccccccccc@{}}
\tblhead{
&\multicolumn{8}{c}{COLORECTAL CANCER}\\
&\multicolumn{2}{c}{ADE vs ECR } &\multicolumn{2}{c}{ADE vs LCR} &\multicolumn{2}{c}{ECR vs LCR} &\multicolumn{2}{c}{CRC vs CTR}}
$(d^+, c^+; d^-, c^-)$ &\multicolumn{2}{c}{(14, 4; 9, 1)}&\multicolumn{2}{c}{(4, 1; 13, 1)}&\multicolumn{2}{c}{(7, 0; 11, 2)}&\multicolumn{2}{c}{(2, 0; 5, 0)}\\
$p^+$ vs $p^-$ &\multicolumn{2}{c}{0.74 vs 0.88} &\multicolumn{2}{c}{0.81 vs 0.92} &\multicolumn{2}{c}{0.83 vs 0.82} &\multicolumn{2}{c}{0.995 vs 1}\\
90\% conf. int.&\multicolumn{2}{c}{[0.64; 0.84] [0.82; 0.94]}&\multicolumn{2}{c}{[0.72; 0.90] [0.87; 0.97]}&\multicolumn{2}{c}{[0.76; 0.90] [0.75; 0.89]}&\multicolumn{2}{c}{[0.99; 1] [1; 1]}
\lastline\\

\tblhead{
&\multicolumn{4}{c}{OVARIAN CANCER (4-3-02)} &\multicolumn{4}{c}{OVARIAN CANCER (8-7-02)}\\
&\multicolumn{4}{c}{Cancer vs Control}&\multicolumn{4}{c}{Cancer vs Control}}
$(d^+, c^+; d^-, c^-)$ &\multicolumn{4}{c}{(7, 2; 5, 1)}&\multicolumn{4}{c}{(6, 0; 4, 0)}\\
$p^+$ vs $p^-$ &\multicolumn{4}{c}{0.93 vs 0.91} &\multicolumn{4}{c}{1 vs 0.98}\\
90\% conf. int. &\multicolumn{4}{c}{[0.89; 0.97] [0.87; 0.95]} &\multicolumn{4}{c}{[1; 1] [0.96; 1]}
\lastline
\end{tabular}}
\end{table}

Indeed, on  colorectal cancer data the discrimination performances  $perf(SA)$ reached by intensive Simulated Annealing for signature discovery  (\cite{Kong2014})  and the performances $perf(MRF)$ reached in the present paper were the following, with error margins of the order of 0.08:\\
for ADE vs ECR: $perf(MRF)= 0.81, perf(SA)= 0.84$, \\
for ADE vs LCR: $perf(MRF)= perf(SA)= 0.86$, \\
for ECR vs LCR: $perf(MRF) =0.82,  perf(SA) = 0.83$,\\ 
for CRC vs CTR: $perf(MRF) = 0.99, perf(SA) =1$. 

For discrimination task Cancer vs Control on the ovarian cancer dataset 4-3-02, our MRF approach  reached  the performance level $perf(MRF) = 0.92$, while the papers  \cite{Assareh2007} and \cite{Kong2014} resp. reported performances $0.92$  and $0.94$. 

For the easier discrimination of Cancer vs Control on the Ovarian Cancer dataset 8-7-02, our MRF approach reached $perf(MRF) = 0.99$, which is equivalent to the  performances of 100\% reported by the four papers \cite{Zhu2003}, \cite{Assareh2007}, \cite{Alexe2004}, \cite{Kong2014}.

\subsection{MRF Discrimination Performance and KL Distance}

\begin{figure}[!h]
\centerline{\includegraphics[width=\textwidth]{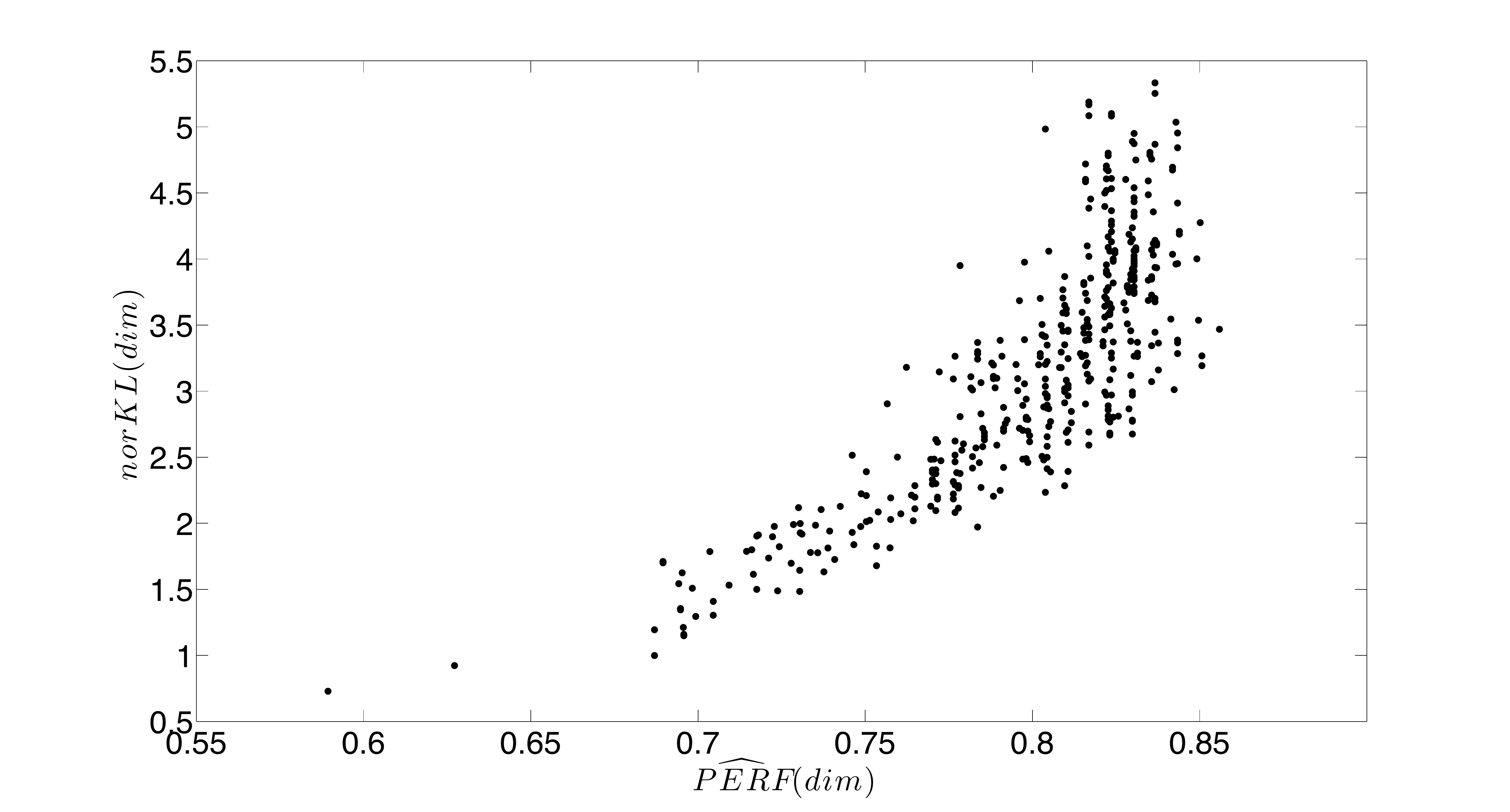}}
\caption{After fitting 2538 pairs $\pi^+_j, \pi^-_j$ of autologistic models to the two groups $G^+ = ECR$  and $ G^- = LCR$ we have computed  the  performance $PERF_j$ of the best separator between $\pi^+_j$ and  $\pi^-_j$, as well as the normalized KL distance $norKL_j$ between $\pi^+_j$ and  $\pi^-_j$. The figure displays the 2538 planar points with coordinates $(PERF_j, norKL_j)$. The graph indicates  a positive correlation between $PERF_j$ and $norKL_j$ particularly for high performances.}\label{perfdisplot}. 
\end{figure}

For each benchmark  discrimination task $G^+$ vs $G^-$, and for each quadruplet $dim =  ( d^+, c^+; d^-, c^- ) $ in $\mathcal{D}(H)$, where $H=15$ or 10, we have fitted as above two autologistic models $\pi^+$ and  $\pi^-$ to the data sets $G^+$ and  $G^-$ and computed the best separator between these two models. After evaluating the discrimination performance $PERF(dim)$  of this separator, we have also systematically computed  the normalized KL distance $norKL(dim)$ between $\pi^+$ and  $\pi^-$ (see Section \ref{KLdistance}). To roughly evaluate  the correlation between $PERF(dim)$ and  $norKL(dim)$ we have then plotted  the planar points $(PERF(dim), norKL(dim))$ on $R^2$. An example of such a plot is given in Figure \ref{perfdisplot} for the discrimination task ECR vs LCR, based on colorectal cancer data. We have observed a fuzzy but positive correlation between  autologistic  discrimination performance $PERF$ and normalized KL distance $norKL$ between pairs autologistic models fitted to $G^+$ and  $G^-$. We have also noted numerically that maximization of $PERF(dim)$ is roughly equivalent to maximization of $norKL(dim)$ over all dimension quadruplets $dim $ in $\mathcal{D}(H)$.

\section{Detailed Implementation for Benchmark Discrimination Task ECR vs LCR}\label{1234}
\subsection{ Sparse Selection of ECR-biomarkers and LCR-biomarkers  }
As just reported, our MRF discrimination approach was successfully tested on cancer data for 6 benchmark discrimination tasks. We now sketch  more detailed implementation steps, but  only for one example, namely the discrimination of  $G^+ = ECR$ versus $G^- = LCR$. This data set involved 154 mass spectra:  80 for  $G^+$ patients and 74 for  $G^-$ patients. Pre-processing of these 154 spectra (see Section \ref{preproc}) yielded a list of 842 reference  biomarkers $B_s$, and hence an initial set $S$ of 842 sites $s$. Binary coding (see Section \ref{bincoding})  of mass spectra then generated two sets $BinG^+$ and $BinG^-$ of binary vectors of length 842, with respective cardinals 80 and 74. 

For each $s \in S$, let $m^+(s)$ and $m^-(s)$ be   the frequencies of the event $x_s = 1$ within $BinG^+$ and $ BinG^-$ respectively. We then focus  only on the set of sites $\hat{S}$ of all $s \in S$ such that  $m^+(s) \geq 20\%$ and $m^-(s) \geq 20\%$, and as above we let $DP(s) = m^+(s)/m^-(s)$. 

For each such quadruplet $dim\in\mathcal{D}(15)$, we select as $G^+$ and  $G^-$ biomarkers  the $d^+$ sites $s \in \hat{S}$ with highest $DP(s)$ and  the $d^-$ sites $s$ with lowest $DP(s)$. As in Section \ref{clique}, we then select $c^+$ pairs of $G^+$ biomarkers, and $c^-$ pairs of $G^-$ biomarkers. The binarized spectra $BinG^+$  and $BinG^-$ are then  restricted to the $d= d^+ + d^-$ sites just selected to define  two sets $BG^+$  and $BG^-$ of binary vectors of length $d$. 
\subsection{ECR vs LCR: Model fitting and Maximization of Performance}
As in Section \ref{numautomodel}, we fit two autologistic distributions $\pi^+$ and $\pi^-$ to $BG^+$  and $BG^-$, to then compute an optimized  separator $\hat{f}$ between $\pi^+$ and $\pi^-$. The empirical performance $\widehat{PERF}(dim)$ of $\hat{f}$ is roughly estimated  as the average percentage of correct discrimination decisions generated by $\hat{f}$ on the training set.

The computation of $\widehat{PERF}(dim)$ is repeated for all $dim \in \mathcal{D}(15)$, and maximization of $PERF(dim)$ over all $dim$ by the accelerated procedure of Section \ref{acceleration} yields the best quadruplet $dim_{opt}$. For  the case $G^+ = ECR$ vs $G^- = LCR$, this  $dim_{opt}$  was $d^+ =7,  c^+ =0; d^-=11, c^- =2$, achieving maximal performance $PERF_{opt}= 0.82$. The associated  separator $\hat{f}_{opt}$ reached percentages of good decisions $p^+= 0.83$ on ECR data and $ p^-=0.82$ on LCR data. 

\begin{table}[!h]
\tblcaption{This table presents the 18 biomarkers retained for optimized discrimination between colorectal cancer groups $G^+ = ECR$ and $G^- = LCR$, as well as their $m/z$ ratios (in Daltons). The columns $m^+(s)$ and $m^-(s) $ display the respective  frequencies of biomarker $B_s$ activation  by the mass spectra in $G^+$ and $G^-$. The top and bottom panels respectively present the  7 biomarkers  with the highest and $m^+(s)/m^-(s)$ ratios  and the 11 biomarkers with the highest $m^-(s)/m^+(s)$ ratios. \label{tab:features}}
{\tabcolsep=4.25pt
\begin{tabular}{@{}cccccccccc@{}}
\tblhead{
\multicolumn{5}{c}{$G^+$ biomarkers} \\
$B_s$ &$m/z$  & $m^+(s)$ & $m^-(s)$ & $ m^-(s) / m^+(s)$}
$B_1$ &1953 &0.21 &0.06 &3.22\\
$B_2$ &1112 &0.29 &0.14 &2.09\\
$B_3$ &8376 &0.55 &0.31 &1.78\\
$B_4$ &6320 &0.43 &0.25 &1.76\\
$B_5$ &2013 &0.24 &0.14 &1.76\\
$B_6$ &921 &0.44 &0.26 &1.72\\
$B_7$ &2558 &0.30 &0.17 &1.71\\
\hline
\multicolumn{5}{c}{$G^-$ biomarkers} \\
$B_s$ &$m/z$  & $m^+(s)$ & $m^-(s)$ & $ m^-(s) / m^+(s)$\\
\hline
$B_8$ &7889 &0.09 &0.26 &2.77\\
$B_{9}$ &2807 &0.15 &0.34 &2.29\\
$B_{10}$ &1032 &0.13 &0.26 &2.03\\
$B_{11}$ &819 &0.17 &0.32 &1.85\\
$B_{12}$ &2143 &0.26 &0.48 &1.81\\
$B_{13}$ &1930 &0.12 &0.21 &1.81\\
$B_{14}$ &2269 &0.12 &0.21 &1.81\\
$B_{15}$ &2815 &0.12 &0.21 &1.81\\
$B_{16}$ &2331 &0.17 &0.31 &1.78\\
$B_{17}$ &1478 &0.15 &0.26 &1.72\\
$B_{18}$ &1210 &0.26 &0.45 &1.72
\lastline
\end{tabular}}
\end{table}
The best autologistic  models $\pi^+_{opt}$ and $\pi^-_{opt}$ associated to $dim_{opt}$ are  based on  a signature of length 18, involving 7 $G^+$ biomarkers $B_1, \ldots, B_7$ and   11 $G^-$ biomarkers $B_8, \ldots, B_{18}$, identified by their  $m/z$ ratios (in Daltons) listed in Table \ref{tab:features}. No clique of $G^+$ biomarkers was retained for  $\pi^+$,  but $\pi^-$  involved 2 cliques of $G^-$ biomarkers  which are listed as well as their $\chi^2$-statistics in Table \ref{tab:JP}. The parameter vectors $\hat\theta^+$ and  $\hat\theta^-$  of $\pi^+_{opt}$ and $\pi^-_{opt}$  generated by MPLE are of dimension 20, and are displayed in two columns  in Table \ref{tab:coeff}, which also gives  the error margins on  parameter estimates. These error margins  define the 90\% confidence intervals computed from  asymptotic normality results (Section \ref{estimerror}). The  zero values displayed for $\hat\theta^+_3$, $\hat\theta^+_4$, $\hat\theta^+_6$, $\hat\theta^-_{12}$, $\hat\theta^-_{18}$ were imposed after a first MPLE evaluation and error  margins computation indicating that these parameters were not significantly different from 0 at the 90\% confidence level.

\begin{table}[!h]
\tblcaption{This table lists the cliques of order 2 for the  best autologistic models $\pi^+$ and $\pi^-$ in the  discrimination task  $G^+=ECR$ vs $G^- =LCR$.  No cliques of $G^+$ biomarkers were retained for $\pi^+$, and 2 cliques of $G^-$ biomarkers were selected for $\pi^-$. The principle is to retain cliques with highest $\chi^2$-statistics. The table displays the 2 selected cliques and the corresponding $\chi^2$-statistics.\label{tab:JP}}
{\tabcolsep=4.25pt
\begin{tabular}{@{}cccccccccc@{}}
\tblhead{
 clique $\{s, t\}$&$m/z$ of $B_s$ &$m/z$ of $B_t$  & $\chi^2(s,t)$}
\{9, 10\}& 2807 & 1032 &7.15\\
\{13,18\}&1930 &1210 &5.45
\lastline
\end{tabular}}
\end{table}

The quality of fit of $\pi^+$ to $ECR$ data and of $\pi^-$ to $LCR$ data were  evaluated as  in Section \ref{QF}, and LCR data and yielded good respective quantile values $46.8\%$ and $51.3\%$ for the log-likelihoods of  the observed ECR and LCR data.

\subsection{ECR vs LCR: Optimized Discriminating Signature and Scores}
The optimized signature discovered for ECR vs LCR discrimination involves 7 ECR-biomarkers $B_1, \ldots , B_ 7$ and 11 LCR-biomarkers $B_8, \ldots , B_{18}$, as well as the 2 pairs of LCR-biomarkers $(B_9, B_{10})$ and $(B_{13}, B_{18})$. As indicated  in Section \ref{signatures}, automatic classification of a mass spectrum $M$ into either ECR or LCR is based on the 18 scores $SCO_1, \ldots , SCO_ {18}$ and the 2 scores $SCO(9,10)$, $SCO(13,18)$. These scores are the coordinates of $ \beta \hat{\theta}^- - \hat{\theta}^+$ and the computed value for $\beta$ was $0.92$. These 20 scores are listed in Table \ref{tab:coeff}. 

\begin{table}[!h]
\tblcaption{This table lists the $G^+$ and $G^-$ biomarkers as well as the only 2 cliques of order 2 retained for the two autologistic models $\pi^+$ and $\pi^-$ enabling  optimal discrimination between $G^+ = ECR$ and $G^- = LCR$. The columns $\hat\theta^+$ and $\hat\theta^-$ display the parameter vectors of $\pi^+$ and $\pi^-$ as estimated by MPLE. For each  coordinate of $\hat\theta^+$ and $\hat\theta^-$, we also give the 90\% confidence intervals  derived from the asymptotic normality results. The zero values displayed for $\hat\theta^+_3$, $\hat\theta^+_4$, $\hat\theta^+_6$, $\hat\theta^-_{12}$, $\hat\theta^-_{18}$ were imposed after a first estimation indicating that these parameters were not significantly different from 0. The two cliques $\{s, t\}$ involve only $G^-$ biomarkers so that the corresponding $\hat\theta^+_{s,t}$ are equal to zero. The table also lists the 20 discrimination scores SCO(s) and $SCO(s, t)$ associated to this signature.\label{tab:coeff}}
{\tabcolsep=4.25pt
\begin{tabular}{@{}cccccccccc@{}}
\tblhead{
\multicolumn{6}{c}{$G^+$ biomarkers} \\
 $B_s$ &$\hat{\theta}^+_s$ &conf. int. of $\hat{\theta}^+_s$ &$\hat{\theta}^-_s$ & conf. int. of $\hat{\theta}^-_s$ &$SCO(s)$}
$B_1$ &1.6 &[1.1; 2.1] &4.3 &[2.5; 6.0] &2.4\\
$B_2$ &1.0 &[0.6; 1.4]&2.3 &[1.6; 3.0] &1.0\\
$B_3$ &0 &-&0.9 &[0.5; 1.3] &0.9\\
$B_4$ &0  &-&1.3 &[0.8; 1.8]&1.2\\
$B_5$ &1.3 &[0.8; 1.8] &2.3 &[1.6; 3.0] &0.8\\
$B_6$ &0 & - &1.2 &[0.7; 1.7]&1.1\\
$B_7$ &1.0 &[0.6; 1.4] &1.9 &[1.3; 2.5] &0.7\\
\hline
\multicolumn{6}{c}{$G^-$ biomarkers} \\
 $B_s$ &$\hat{\theta}^+_s$ &conf. int. of $\hat{\theta}^+_s$ &$\hat{\theta}^-_s$ &conf. int. of $\hat{\theta}^-_s$ &$SCO(s)$\\
\hline
$B_8$&2.9 &[2.1; 3.7]&1.2&[0.7; 1.7] &-1.8\\
$B_9$&2.1 &[1.5; 2.7]&1.1&[0.6; 1.6]&-1.0\\
$B_{10}$ &2.3 &[1.6; 3.0] &1.8 &[1.1; 2.5]&-0.7\\
$B_{11}$&1.8 &[1.3; 2.3]&0.9 &[0.5; 1.3]&-1.0\\
$B_{12}$&1.2 &[0.8; 1.6]&0 &-&-1.2\\
$B_{13}$ &2.5 &[1.8; 3.2] &1.0 &[0.4; 1.6]&-1.6\\
$B_{14}$&2.5 &[1.8; 3.2]&1.5&[1.0; 2.0]&-1.1\\
$B_{15}$&2.5 &[1.8; 3.2]&1.5&[1.0; 2.0]&-1.1\\
$B_{16}$ &1.8 &[1.3; 2.3]&0.9 &[0.5; 1.3]&-1.0\\
$B_{17}$&2.1&[1.5; 2.7]&1.2&[0.7; 1.7]&-1.0\\
$B_{18}$&1.2 &[0.8; 1.6]&0&-&-1.2\\
\hline
\multicolumn{6}{c}{$G^-$ cliques} \\
$\{s, t\}$&$\hat{\theta}^+_{s, t}$ &conf. int. of $\hat{\theta}^+_{s,t}$ &$\hat{\theta}^-_{s, t}$ & conf. int. of $\hat{\theta}^-_{s,t}$ &$SCO(s, t)$\\
\hline
\{9, 10\} &0 &-&-1.5 &[-2.5; -0.5]&-1.4\\
\{13, 18\} & 0 &-&1.7 &[0.4; 3.0]&1.6
\lastline
\end{tabular}}
\end{table}

For any mass spectrum $M$ and any $s = 1, \ldots , 18$, recall that we have set $x_s(M)=1$ when  the biomarker $B_s$ is activated by $M$, and $x_s(M)=0$ otherwise. The total score $TSCO(M)$ is here given by 
\begin{align*}
TSCO(M) &= 0.37 +  x_1(M) SCO(1) + \ldots +  x_{18}(M) SCO(18)  \\
&+ x_9(M) x_{10}(M) SCO(9,10) + x_{13}(M) x_{18}(M) SCO(13,18)
\end{align*}
Then $M$ is classified as belonging to the cancer group $ECR$ if $TSCO(M) > 0$ and to the group $LCR$ otherwise.

\section{Conclusion}

Mass spectrometry is a  promising approach for biomarker-based early cancer detection. By algorithmic analysis of  mass spectra acquired by MALDI-TOF or SELDI-TOF techniques, selected sets of peptides strongly linked to specific cancer groups have been used for automated   cancer stage classification. To further help incorporating these approaches in  reliable clinical protocols, a key step is to discover  interpretable ``biomarker signatures ``characterizing various cancer types and/or cancer stages.

Machine learning algorithms, such as decision trees, support vector machines, artificial neural networks, etc., have been tested on mass spectra  data sets acquired from cancer patients, and have often been proposed as efficient methods to discriminate between groups of mass spectra. But  these techniques tend to generate   ``black-box" results which often lack direct biological interpretability. 

Our main focus was to rigorously fit parameterized stochastic models to mass spectra datasets acquired by MALDI or SELDI techniques, in  order to design efficient signature discovery algorithms leading to interpretable signatures combining the discriminating power of well selected \textit{small} groups of biomarkers.

In this paper, the pattern variations observable in any supposedly ``homogeneous " data set acquired by mass spectrometry have been systematically  modeled by  binary Markov Random Fields (MRF). After fairly classical pre-processing of any given group G of mass spectra, we generate a reference list of several hundreds to several thousands of strong spectral ``peaks", viewed as potential key biomarkers to characterize the group G. We then code each mass spectrum M by a long binary vector  listing  the binary status (presence/absence in M) of each reference biomarker. Gibbs distributions, as used in our work, are efficient stochastic models to study the spatial dependency of coordinates for high dimensional binary vectors viewed as realizations of Markov Random Fields. We have focused our study on autologistic models - a type of Gibbs distributions involving only paiwise interactions between binary sites. Automatic classification of new mass spectra into two distinct mass spectra datasets $G^+$ and $G^-$ can then be reduced to computing  the best  classifier discriminating  two autologistic models.

Based on this theoretical point of view and associated algorithms, we have proposed a systematic approach to discover explicit and highly discriminating biomarker signatures, enabling efficient discrimination between distinct homogeneous groups of binary coded mass spectra. To construct autologistic models of random peaks patterns variations among mass spectra,   stochastic model fitting to data was implemented by Maximum Pseudo-Likelihood Estimation (MPLE), and  achieved a good quality of fit to mass spectra data in all our cancer data sets  studies.

We have successfully tested our innovative signature discovery  algorithms on a new experimental set of MALDI-TOF mass spectra acquired from patients at three stages of colorectal cancer, as well as on two previously published data sets of SELDI-TOF mass spectra acquired from ovarian cancer patients. Final  performance levels are computed by leave-one-out cross validation. The performance of our algorithm is good in all these concrete cases and compared quite favorably with  performance levels reported in previous publications. The clear concrete advantage of our optimized signature discovery technique is that it provides a biologically interpretable signature, involving  only a small number of   key biomarkers identified by their m/z ratios, and  where each biomarker is weighted by an explicit numerical score.

\section*{Acknowledgments}
The authors thank M. Agostini  and C. Bedin at the 1st Surgical Clinic, Dept of Surgical, Oncological and Gastroenterological Sciences (University of Padova) for providing plasma samples from colorectal cancer patients; we also thank A. Bouamrani, E. Tasciotti and M. Ferrari from The Methodist Hospital Research Institute of Houston (Dept. of Nanomedicine) for providing us with  the  MALDI-TOF mass spectra acquired from these plasma samples. We have also benefitted from multiple interesting  and informative discussions with the five specialists just mentioned, as well as with C. Gupta (Mathematics, University of Houston),  in the framework of the collaboration which led to the joint publication  \cite{Kong2014}.

{\it Conflict of Interest}: None declared.

\section*{Appendix}
\setcounter{proposition}{0}
\setcounter{section}{0}
\renewcommand{\thesection}{\Alph{section}}

This mathematical appendix outlines concise and self-contained  proofs of several theoretical results stated and used in this paper. Indeed, in  the literature on MRFs, Gibbs distributions, and MPLE estimation on finite configuration spaces,  similar results have been used at least  implicitly, but elements of proof are often absent, or linked to very  different contexts, or imbedded within proofs of other theorems. 

In all the following theorems and proofs, $\mathcal{B}$ is a space of binary vectors of fixed length $L$.

\section{Existence and Uniqueness of Maximum Pseudo-Likelihood Estimators}\label{MPLEexist}
Consider as in Section \ref{models}, and with the same notations, an autologistic distribution $\pi_{\theta}$ parameterized by the unknown vector $\theta \in \Theta$. Given $n$ observed random binary vectors generated by $\pi_\theta$, the MPLE estimator  $\hat{\theta}_n$ of $\theta$ is computed by maximizing in $T \in \Theta$ the empirical pseudo-likelihood $ \hat {g}(T,n) $ as defined by \eqref{MPLEobj}. Existence and fast computability of the MPLE  is due to the concavity of  $ \hat {g}(T,n) $ and of its limit $g_\theta (T)$ as $n \to \infty$, as we now prove.

\begin{proposition}  For each $\theta \in \Theta$, the mean log pseudo-likelihood $g_\theta (T)$ defined by equation \eqref{gtheta} is  a strictly concave function of $T \in \Theta$, and reaches its maximum in $T$ at the unique point $T= \theta$. Moreover, the empirical pseudo-likelihood $ \hat {g}(T,n) $ is also concave in $T \in \Theta$, and becomes almost surely strictly concave as $n \to \infty$.           
\end{proposition}
\textbf{Proof:} (Notations of Section \ref{pseudo} and \ref{computeMPLE}). The functions $A(s,x), g(z), h(z), LPL_{s,x}(T)$ were defined by equation \eqref{Asx}, \eqref{g&h}, and \eqref{LPLsxT}. For any binary number $x_s$, the function $\;\phi(z) = x_s \, z + \log(g(z)) \; $ then verifies 
$$
\phi'(z) = \; x_s- h(z) \;; \quad\quad \phi''(z)= \; - e^z/(1+e^z)^2 < 0.
$$
Since $LPL_{s,x}(T) = \phi(<A(s,x),T>)$, the gradient in $T$ of $LPL_{s,x}(T)$ is the vector 
\begin{equation} \label{gradLPL}
\partial_T LPL_{s,x}(T) = \phi'(<A(s,x),T>)\; A(s,x)  =\left[ \, x _s - h( <A(s,x),T> ) \, \right] A(s,x).
\end{equation}
The Hessian $HLPL(s,x,T) $ of $LPL_{s,x}(T)$ with resp. to $T$ is hence the quadratic form defined on all vectors $t \in \Theta$ by 
$$
t^* \, HLPL(s,x,T) \, t = \phi''( <A(s,x),T> )\; <A(s,x),t>^2 
$$
so that 
$$ t^* \, HLPL(s,x,T) \, t  = \frac{- <A(s,x),t>^2 \; e^{<A(s,x),T>}}{(1+e^{<A(s,x),T>})^2} \leq 0.
$$
The Hessian $Hg_\theta(T)$ of $g_\theta(T)$ with respect to $T$ is hence the quadratic form defined on all vectors $t \in \Theta$ by 
\[
t^* \, Hg_\theta(T) \, t = 
\sum_s \sum_x\left[\, \pi_\theta(x) \phi''(<A(s,x),T>)<A(s,x),t>^2 \, \right] \leq 0.
\]
The quadratic form $t^* \, Hg_\theta(T) \, t $ in $t \in \Theta$ is thus clearly non positive. Since $\pi_\theta(x)> 0$ and $ \phi''(z) <0 $, this quadratic form takes the value 0 for some $t \in \Theta$ if and only if $<A(s,x),t> = 0 $ for all $s \in S$ and all binary configurations $x$. But the conditions $t_s+\sum_{s < r} t_{sr} x_r = 0 $ for all indices $s$ and all binary vectors $x$ are easily shown to force $t= 0$. 

The Hessian of $g_\theta(T) $ is hence negative definite for all $ T $, and thus the function $g_\theta(T)$ is strictly concave for all $T \in R^k$. Since $g_\theta$ is obviously bounded, it has a unique maximum in $T$, and this maximum is reached at the unique vector $\hat{T}$ which cancels the gradient $\partial_T g_\theta(T)$ of $g_\theta(T)$. So to conclude that $\hat{T}= \theta$, we now prove that $\partial_T g_\theta(\theta ) = 0$.

Due to equations \eqref{gradLPL} and \eqref{aA}, for all indices $s$ and $s < r $, and for all binary configurations $x$, the partial derivatives of $LPL(x,T)$ are given by 
\begin{align}
&\frac{\partial LPL(x,T) }{\partial T_s} = - x_s + h( < A(s,x),\theta > ),   \label{Ds} \\
&\frac{\partial LPL(x,T) }{\partial T_{s r}} = - 2 x_s x_r + x_r \, h( < A(s,x),\theta > )+  x_s \,  h( < A(s,x),\theta > ). \label{Dsr}
\end{align} 
Letting $T=\theta$ in the conditional specifications formulas given above, we get, for a random configuration $X$ with distribution $\pi_\theta$, 
$$
P_\theta ( X_s=1 \mid X_{S-s} ) = h( < A(s,X),\theta > )
$$
and hence for all indices $s$ and $s < r $ one has 
\begin{align*} 
&E_\theta ( h( < A(s,X),\theta > ) )= E_\theta ( P_\theta ( X_s=1 \mid X_{S-s} ) = P_\theta ( X_s=1 ) =  E_\theta ( X_s), \\
&E_\theta ( X_r \, h( < A(s,X),\theta > ))= E_\theta ( X_r P_\theta ( X_s=1 \mid X_{S-s} ) ) =  E_\theta ( X_r X_s). 
\end{align*}
Due to equations \eqref{Ds}, \eqref{Dsr}, these identities prove that for all indices $s$ and $s<r$
\[
\frac{\partial g_\theta(\theta) }{\partial T_{s}} = \; 
E_\theta \;(\; \frac{\partial LPL( X, \theta) }{\partial T_{s}}\; ) = 0 \quad \textit{and}\quad
\frac{\partial g_\theta(\theta) }{\partial T_{s r}} = \; E_\theta \;(\; \frac{\partial LPL( X, \theta) }{\partial T_{s r}}\; ) = 0.
\]
We have thus shown that the gradient $\partial_T g_\theta(T)$ is equal to $0$ when $ T= \theta$. This concludes the proof.
\section{Asymptotic Normality of MPLE}\label{asymnorm}
\begin{proposition}  For any autologistic distribution $\pi_\theta$ on the finite set of binary configurations $\mathcal{B}$, the MPLE estimators $\hat{\theta}(n)$ of  $\theta$ are asymptotically consistent as the number of observations $ n \to \infty$. The normalized vectors of estimation errors $\sqrt{n} \; (\hat{\theta}(n) - \theta)$ are asymptotically Gaussian with mean zero and  covariance matrix 
\[
\Gamma(\theta)= \mathcal{H}(\theta)^{-1} \Sigma(\theta) \mathcal{H}(\theta)^{-1},
\] 
where $\mathcal{H}(\theta)$ is the Hessian of $g_\theta (T)$ at $T=\theta$, and the symmetric positive definite matrix $ \Sigma(\theta) $ is determined  by equations \eqref{SigmaT} (see below).
\end{proposition}

\textbf{Proof:}  (Notations of Section \ref{pseudo} and \ref{computeMPLE}). We only outline the main technical steps, since our  approach  is similar to the "contrast function" analysis applied in the last chapter of \cite{Azencott1985}. 

For all fixed parameter vectors $\theta \in \Theta $ the contrast function $g_\theta(T)$ is strictly concave in $T$, as seen above, and the law of large numbers gives the almost sure limit $ \lim_{n\rightarrow \infty} \hat {g}(T,n) = g_\theta(T)$.

Asymptotic consistency of $\hat{\theta}(n)$ is then derived as in \cite{Azencott1985}. 

Let $\hat{G}(T,n)$  and $\hat{H}(T,n)$  be the  gradient and the Hessian  of  $\hat {g}(T,n)$
as defined by equations \eqref{hatgradG} and   \eqref{hathessH}. Apply first the central limit theorem to assert that the random vector 
$$
Y(T,n) = \sqrt{n} \; ( \hat{G}(T,n) - \partial_T g_\theta(T) )
$$
is asymptotically Gaussian with mean zero and covariance matrix 
\begin{equation}\label{SigmaT}
\Sigma(T)= E_\theta ( \partial_T LPL(X,T) \; \partial_T LPL(X,T)^* ),
\end{equation}
where $X$ is a random configuration with distribution $\pi_\theta$.\\
Let $ \;\Sigma_{s,j} \, , \Sigma_{st,ij} \, , \Sigma_{s,ij} \; $, with $s,t,i,j \in S$ and $ s < t , \; i < j $, denote the elements of the symmetric matrix $\Sigma(T)$. \\
Let $W_s= h( < A(s,X), T > )$, and apply formulas \eqref{Ds} and \eqref{Dsr} to obtain 
\begin{align*}
&\Sigma_{s,i}(T) = E_\theta [ \; (-X_s + W_s)(-X_i + W_i) \; ],  \\
&\Sigma_{st,ij}(T) = 
E_\theta [ \; (-2 X_s X_t +  X_t W_s +   X_s W_t) (-2 X_i X_j +  X_j W_i + X_i W_j)\; ], \\
&\Sigma_{s,ij} (T) = 
E_\theta [ \; (-X_s +  W_s ) (-2 X_i X_j +  X_j W_i + X_i W_j) \; ].
\end{align*}
For any given $T$, these expressions can easily be used to compute good numerical estimates of $\Sigma(T)$ by simulated Gibbs sampling of $\pi_\theta$.

For $T$ close to $\theta$, and since $\partial_T g_\theta(\theta)  = 0 $ as seen above, we have by Taylor's formula, using the Hessian $Hg_\theta(T)$ of $g_\theta(T)$
$$
\partial_T g_\theta(T) = \partial_T g_\theta(T) - \partial_T g_\theta(\theta) ) \sim 
Hg_\theta(\theta)(T- \theta).
$$
Since the MPLE estimator $\hat{\theta}_n$ verifies $\hat{G}(\hat{\theta}(n), n)= 0$, we have 
$$
Y(\hat{\theta}_n, n) = \sqrt{n} \; ( \hat{G}(\hat{\theta}_n,n) - \partial_T g_\theta(\hat{\theta}_n) )
= - \sqrt{n} \; \partial_T g_\theta(\hat{\theta}_n) ).
$$
The last two relations give then the approximation 
$$
- Y(\hat{\theta}_n, n) \sim \sqrt{n} \; Hg_\theta(\theta) \, ( \hat{\theta}_n - \theta).
$$
Since for $n$ large, $\hat{\theta}_n$ is close to $\theta $ in probability, we see that with high probability $Y(\hat{\theta}_n, n)$ is close to $Y(\theta, n)$. We conclude that for $n$ large, one has with high probability the approximation 
$$
- Y(\theta, n) \sim \sqrt{n} \; Hg_\theta(\theta) ( \hat{\theta}_n - \theta).
$$
Since the Hessian $\mathcal{H}(\theta)= Hg_\theta(\theta)$ is an invertible matrix, the random vector 
$$
\sqrt{n} \; ( \hat{\theta}_n - \theta) \sim - \mathcal{H} (\theta)^{-1} Y(\theta, n)
$$
must become, for $n$ large, approximately Gaussian with mean zero and covariance matrix 
$\mathcal{H} (\theta)^{-1}\Sigma(\theta) \mathcal{H} (\theta)^{-1}$.
\section{Estimated optimal separator } \label{estisep}
Fix two autologistic distributions  $\pi^+, \pi^- $ on $\mathcal{B}$ parameterized by $\theta^+, \theta^-$ in $\Theta$.  Let $\hat{\theta}_n^+$ and $\hat{\theta}_n^-$ be the  MPLE estimators of $\theta^+, \theta^-$, computed   from two separate samples of  $n$   configurations resp. generated by $\pi^+$ and $\pi^-$.  The  optimal separator $f(x) = < u,\, U(x) > +\, a$ between  $\pi^+$ and $\pi^-$, and its  natural estimator $ \hat{f}(x) = < \hat{u},\, U(x) > +\, \hat{a}$   are given by formulas \eqref{f(x)} and \eqref{hatf(x)}.

\begin{proposition}  In the preceding situation, as $n \to \infty$, the normalized error vector $\sqrt{n} ( \hat{u} - u )$ is asymptotically Gaussian with mean zero and covariance matrix $Cov = \Gamma(\theta^+) + \Gamma(\theta^-) $, where the matrices $\Gamma(\theta)$ have been computed in Proposition \ref{propasym}. Moreover, $\sqrt{n} \, ( \hat{a} - a )$ is also asymptotically Gaussian with mean zero and asymptotic variance $\var(\theta^+, \theta^-)$ verifying 
\begin{equation}\label{var}
\var(\theta^+, \theta^-) \leq 
\frac{L(L+1)}{2} \times trace \, \left[ \, \Gamma(\theta^+) + \, \Gamma(\theta^-) \right].
\end{equation}
\end{proposition}
\textbf{Proof:} For  $\theta \in \Theta$, let $\varepsilon_n(\theta) = \sqrt{n} \, ( \hat{\theta}_n - \theta )$ be the vector of normalized estimation errors. The independent vectors $\varepsilon_n(\theta^+) $ and $\varepsilon_n(\theta^-) $ are both asymptotically Gaussian with mean zero and covariance matrices $ \Gamma(\theta^+)$, $ \Gamma(\theta^-)$ computable via Proposition \ref{propasym}. Hence the vector $\sqrt{n} ( \hat{u} - u )$ is asymptotically Gaussian with mean zero and covariance matrix $Cov = \Gamma(\theta^+) + \Gamma(\theta^-) $.

The partition function $Z(\theta) = \sum_{x \in \mathcal{B}} \; \exp(- \theta U(x))$ of $\pi_\theta$ is a smooth positive function of $\theta \in \Theta$. The vector valued function $ \psi(\theta) = \left[ \, \theta, z(\theta) \, \right] $ where $z(\theta) = \log(Z(\theta))$ is hence also a smooth in $\theta$. As is  well known (see for instance \cite{Azencott1985}), smooth functions preserve asymptotic normality. Hence  since $\varepsilon_n(\theta) $ is asymptotically Gaussian with mean zero and covariance matrix $\Gamma(\theta)$, we conclude that 
$\sqrt{n} \, ( \psi(\hat{\theta}_n) - \psi(\theta) )$ must be asymptotically Gaussian with mean zero and covariance matrix $\psi'(\theta)^* \Gamma(\theta) \psi'(\theta) $, where the gradient $\psi'(\theta)$ is given by 
\[
\psi'(\theta) v = \left[ \, v , v^* z'(\theta) \, \right] \quad \textit{for all vectors} \quad v \in \Theta.
\]
In particular, $\sqrt{n} \, ( z(\hat{\theta}_n) - z(\theta) )$ is asymptotically Gaussian with mean zero and variance 
\[
\tau(\theta) = z'(\theta)^* \Gamma(\theta) z'(\theta),
\]
where the column vector $z'(\theta)$ is the gradient of $z(\theta)$. 

By a known trace identity, one has then the upper bound 
\[
\tau(\theta) \leq || \, z'(\theta) \, ||^2 \; \textit{trace} \, \left[ \, \Gamma(\theta) \, \right]. 
\]
The gradient  of $Z(\theta)$ is given by 
\[
Z'(\theta)=  - \sum_{x \in \mathcal{B}} \;  U(x) \, \exp( - < \theta, U(x) >). 
\] 
Since $U(x)$ has dimension $ k= L(L+1)/2$, one has $|| \, U(x) \, || \leq \sqrt{k} $, and hence  
$|| \, Z'(\theta) \, || \leq \sqrt{k} Z(\theta)$.

This yields immediately $ || \, z'(\theta) \, || = || \, Z'(\theta)/Z(\theta) \, || \leq \sqrt{k} $ 
and thus finally 
\[
\tau(\theta) \leq k \times trace \, \left[ \, \Gamma(\theta) \, \right]. 
\]
Applying this result to $\theta^+$ and $\theta^-$ shows that $\sqrt{n} \, ( \hat{a} - a )$ is asymptotically Gaussian with mean zero and variance verifying inequality \eqref{var}. This concludes the proof.

\begin{proposition}  On the set $\mathcal{B}$ of binary vectors $x$, consider the two  separators 
\[
f(x)= < u, U(x) > + a \; ; \quad \phi(x)= < \eta, U(x) > + \alpha 
\]
parameterized by $u, \eta \in {\Theta}$ and  $a, \alpha \in R$.
Fix any autologistic  distribution $\pi_\theta$ on $\mathcal{B}$. Let $p(f)$ and $p(\phi)$ be the two probabilities 
\[
p(f) = \pi_\theta [ \, x \mid f(x) >0 \, ]\quad
\textit{and}\quad p(\phi) = \pi_\theta [ \, x \mid \phi(x) >0 \, ].
\]
Recall that $k = L(L+1)/2$. Fix $u,a, \theta$ and assume that $0 < p(f) < 1$. For any arbitrary small percentage $0\% \leq \gamma < 100\%$, there is a strictly positive number $q$ depending only on $\theta, u, a, \gamma$ such that for any pair $(\eta, \alpha)$ verifying
\begin{equation} \label{cond}
|| \eta -u || < \frac{q}{2\sqrt{k}} \quad \textit{and} \quad | \, \alpha - a \, | < q/2,
\end{equation} 
one must then have $ | \, p(f) - p(\phi) \, | \leq \gamma$.
\end{proposition} 

\textbf{Proof:} There exists a \textit{strictly positive} number 
$q = q(\theta, u, a, \gamma)$ such that 
\begin{align*}
&\pi_\theta \left\lbrace \; x \in \mathcal{B} \; \mid \; 0 < f(x) < q \; \right\rbrace \; \leq 
\; \gamma \, p(f), \\
&\pi_\theta \left\lbrace \; x \in \mathcal{B} \; \mid \; - q < f(x) < 0 \; \right\rbrace \; \leq 
\; \gamma \, ( 1 - p(f) ). 
\end{align*}
Indeed, since $\mathcal{B}$ is finite, this is already true for $\gamma = 0$, and hence a fortiori for any $\gamma \geq 0$. But for any given $\gamma$, we will systematically select $q$ as the largest number verifying the preceding two inequalities.

Write  $F= \left\lbrace \, x \in \mathcal{B} \mid f(x) >0 \, \right\rbrace$ and 
$\Phi= \left\lbrace \, x \in \mathcal{B} \mid \phi(x) >0 \, \right\rbrace $ so that  $p(f) = \pi_\theta(F)$ and $ p(\phi) = \pi_\theta(\Phi)$. By construction of $q$, one has then the inequalities 
\begin{align} 
&\pi_\theta( F - \Phi ) \leq \gamma \, p(f)
 + \pi_{\theta} \; \left\lbrace \; x \in \mathcal{B}  \mid 
f(x) \geq q > 0 \;\; \textit{and} \;\; \phi(x) <0 \; \right\rbrace, \label{ineq} \\
&\pi_\theta( \Phi - F ) \leq \gamma \, ( 1 - p(f) )
 + \pi_{\theta} \; \left\lbrace \; x \in \mathcal{B} \mid 
f(x) \leq - q < 0 \;\; \textit{and} \;\; \phi(x) > 0 \; \right\rbrace. \label{ineqbar}
\end{align}
The two joint conditions $f(x) \geq q > 0 $ and $\phi(x) <0$ imply the elementary inequalities 
\[
q \leq f(x) - \phi(x) = < u- \eta,\,  U(x) > +\, a - \alpha\leq \; \sqrt{k} || u -\eta || 
+ | \, a-\alpha \, |,
\]
where $|| . ||$ is the  norm in $R^k$ and one uses the bound $||U(x)|| \leq \sqrt{k}$. 
Similarly the joint conditions $f(x) \leq - q < 0 $ and $\phi(x) > 0$ imply 
\begin{align*}
&q \leq - f(x) + \phi(x) = < - u + \eta,\\
&U(x) >   + \alpha -a  \; \leq \; \sqrt{k} || u -\eta || 
+ | \, (a-\alpha) \, |.
\end{align*}
Hence for any pair $(\eta, \alpha)$ verifying the conditions \eqref{cond}, we see that the 2nd terms in both inequations \eqref{ineq} and \eqref{ineqbar} must be zero, which implies 
\[
\pi_\theta( F - \Phi ) \leq \gamma \, p(f) \quad \textit{and} \quad \pi_\theta( \Phi - F) ) \leq \gamma \, ( 1-p(f) ).
\]
Since one clearly has
\[
| \, p(f) - p(\phi) \, | \leq \pi_\theta( F - \Phi ) + \pi_\theta( \Phi - F) ),
\]
we conclude that $ | \, p(f) - p(\phi) \, | \leq \gamma $.

\section{Control of discrimination errors due to separator estimation}\label{ctrerror}

\begin{proposition}  Fix $\theta^+, \theta^- \in \Theta$ and let $\pi^+= \pi_{\theta^+}$ and $\pi^- = \pi_{\theta^-}$. Let $f(x) = u U(x) + a$ be the optimal separator between $\pi^+$ and $\pi^-$, given by equation \eqref{f(x)}. From two random samples of $n$ configurations resp. generated by $\pi^+$ and $\pi^-$, one computes by equation \eqref{hatf(x)} the estimator $\hat{f}(x) = \hat{u} U(x) + \hat{a}$ of $f(x)$. Let $( p^+, p^- )$ and $( \hat{p}^+, \hat{p}^- )$ be the probabilities of correct decisions resp. achievable by the separators $f$ and $\hat{f}$ for discrimination between $\pi^+$ and $\pi^-$. Assume that $0 < p^+ < 1$ and $0 < p^- < 1$. 

Let $\gamma$ and $\kappa$ be two arbitrary small numbers verifying $0 \leq \gamma < 1$
and $0 < \kappa < 1$. Then one can find $N$ such that for $n > N$ one has 
\[
P (\; | \, \hat{p}^+ - p^+ \, | \leq \gamma \;\; \textit{and} \;\;
| \, \hat{p}^- - p^- \, | \leq \gamma \; ) \; \geq \; 1 - \kappa. 
\]
A practical estimate of $N$ can be computed using  formula \eqref{N} below.
\end{proposition} 
\textbf{Proof:} Let $q>0$ be the largest strictly positive number verifying
\begin{align*} 
&\pi^+ \left\lbrace \; x \in \mathcal{B} \; \mid \; 0 < f(x) < q \; \right\rbrace \; \leq 
\; \gamma \, p^+, \\
&\pi^+ \left\lbrace \; x \in \mathcal{B} \; \mid \; - q < f(x) < 0 \; \right\rbrace \; \leq 
\; \gamma \, ( 1 - p^+ ). 
\end{align*}
Fix a number $A>0$, which will be explicitly selected below. For $\; n > 4 k A^2/q^2 \;$ we have the elementary inequalities 
\[
P(\; || \, \hat{u} - u \, || \geq \frac{q}{2\sqrt{k}} \, ) = 
P(\; \sqrt{n} \, || \, \hat{u} - u \, || \geq \sqrt{n} \, \frac{q}{2\sqrt{k}} \, )\leq P(\; \sqrt{n} \, || \, \hat{u} - u \, || \geq A \, ).
\]
Let $V \in \Theta$ be a random Gaussian vector with mean zero and covariance matrix $Cov$. The asymptotic normality results proved above now imply 
\begin{equation}  
\lim_{n \to \infty} P(\; || \, \hat{u} - u \, || \geq \frac{q}{2\sqrt{k}} \, ) \leq \lim_{n \to \infty} P(\; \sqrt{n} \, || \, \hat{u} - u \, || \geq A \, )= P(\; ||V|| \geq A \;).\label{bound}
\end{equation} 
Let $\lambda_1 \geq \ldots \geq \lambda_k >0$ be the eigenvalues of the covariance matrix $Cov$. By diagonalization of $Cov$, one finds a $(k \times k)$ orthogonal matrix $M$ such that the coordinates of $M V$ become $ \left[ \; \sqrt{\lambda_1} \, H_1 , \ldots , \sqrt{\lambda_k} \, H_k \; \right]$, where the $H_j$ are independent standard Gaussian random variables. Since $|| MV || = || V ||$, we can then write 
\[
P(\; ||V||^2 \geq A^2 \;)= P(\; \sum_{j=1}^k \lambda_j H_j ^2 \geq A^2 \;)\leq P(\; \lambda_1 \sum_{j=1}^k \; H_j^2 \geq A^2 \;).
\]
Fix any small $\kappa > 0$. Since $ \sum_{j=1}^k \; H_j^2 $ has a $\chi^2$ distribution with $k$ degrees of freedom, let  $Q(\kappa)$ be the $100 \times (1-\kappa)\%$ percentile of the  $\chi^2_k$ distribution. Now select and fix $A^2 = Q(\kappa) \lambda_1$ in the last inequation to derive 
\[
P(\; ||V|| \geq A \;) \leq \kappa.
\]
After imposing $\; n > 2k A^2/q^2 = 4k \lambda_1 \, Q(\kappa) /q^2 \;$, inequation \eqref{bound} now implies
\begin{equation} \label{1}
\lim_{n \to \infty} P(\; || \, \hat{u} - u \, || \geq \frac{q}{2\sqrt{k}} \, ) \leq \kappa.
\end{equation}
Let $R(\kappa)$ be the $100 \times (1-\kappa)\%$ percentile of the standard Gaussian distribution. After imposing $n > 4 R(\kappa)/q^2$, the asymptotic normality of $\sqrt{n} \,(\hat{a} -a)$ proved above immediately shows, that 
\begin{equation} \label{2}
\lim_{n \to \infty} P(\; | \, \hat{a} - a \, | \geq \frac{q}{2} \, ) \leq \kappa.
\end{equation}
Combining inequations \eqref{1} and \eqref{2} immediately yields
\begin{equation} \label{3}
\lim_{n \to \infty} P(\; || \, \hat{u} - u \, || < \frac{q}{2\sqrt{k}} \quad \textit{and} \quad | \, \hat{a} - a \, | < \frac{q}{2} \; ) \; \geq 1 - 2 \kappa.
\end{equation}
Let $\hat{p}^+ = \pi^+ \, [ \, x \, \mid \, \hat{f}(x) >0 \, ]$ be the probability of good decisions under $\pi^+$ based on the estimated separator $\hat{f}(x)$. In view of inequation \eqref{3}, we can now apply the Proposition \ref{properror} to conclude that
\[
\lim_{n \to \infty} P (\; | \, \hat{p}^+ - p^+ \, | \leq \gamma \; ) \; \geq 1 - 2 \kappa.
\]

In fact the asymptotic normality results used above are practically accurate in most concrete cases as soon as $n > 50$. So we see that when the separator $\hat{f}$ is estimated from $n$ observations generated by $\pi^+$ and $n > N$ observations generated by $\pi^-$, with 
\begin{equation} \label{N}
N =  \max \; \left\lbrace \; 50 , n > 4 R(\kappa)/q^2 , 4k \lambda_1 \, Q(\kappa) /q^2 \right\rbrace, 
\end{equation}
we can essentially assert that $\; P(\; | \, \hat{p}^+ - p^+ \, | \leq
\gamma \; ) \geq 1 - 2 \kappa \;$.\\
Now to prove that 
\[
\lim_{n \to \infty} P ( \; | \, \hat{p}^- - p^- \, | \leq \gamma \; ) \; \geq 1 - 2 \kappa,
\]
one can define a new $q >0 $ as the largest number verifying the two equations 
\begin{align*} 
&\pi^- \left\lbrace \; x \in \mathcal{B} \; \mid \; 0 < f(x) < q \; \right\rbrace \; \leq 
\; \gamma \, p^-, \\
&\pi^- \left\lbrace \; x \in \mathcal{B} \; \mid \; - q < f(x) < 0 \; \right\rbrace \; \leq 
\; \gamma \, ( 1 - p^- ), 
\end{align*}
and then apply the same arguments as above. This achieves the proof.

\bibliographystyle{biorefs}
\bibliography{mybib}

\end{document}